\pdfoutput=1
\documentclass[11pt]{article}

\usepackage[preprint]{acl}

\usepackage{times}
\usepackage{latexsym}

\usepackage[T1]{fontenc}
\usepackage[utf8]{inputenc}

\usepackage{microtype}

\usepackage{microtype}
\usepackage{booktabs}       % professional-quality tables
\usepackage{nicefrac}       % compact symbols for 1/2, etc.
\usepackage{xcolor}         % colors
\usepackage{hyperref}
\usepackage{url}
\usepackage[oldenum]{paralist}
\usepackage{amsmath}
\usepackage{amsfonts}
\usepackage{amssymb,stmaryrd}
\usepackage{verbatim}
\usepackage{subfigure}
\usepackage[linesnumbered,lined,boxed,commentsnumbered,ruled,vlined]{algorithm2e}
\usepackage{algorithmic}
\usepackage{tikz}
\usepackage{pgfplots}
\pgfplotsset{compat=1.18} 
\usepackage{array}
\usepackage{multirow}
\usepackage{makecell}
\usepackage{enumerate}
\usepackage{enumitem}
\usepackage{bm}
\usepackage{comment}
\usepackage{soul}
\usepackage{footmisc}
\usepackage[normalem]{ulem}
\usepackage{minibox}
\usepackage{framed}
\graphicspath{{figures/}}
\usetikzlibrary{positioning}
\usepackage{pgfgantt}
\allowdisplaybreaks           		
\usepackage{graphicx}				
\usepackage{amssymb}
\usepackage{arydshln}
\usepackage{bm}
\usepackage{tcolorbox}

\usepackage{inconsolata}

\def \OURS {{TRACE}}
\newcommand{\jy}[1]{\textcolor{black}{#1}}

\title{TRACE the Evidence: Constructing Knowledge-Grounded Reasoning Chains for Retrieval-Augmented Generation}

\author{
Jinyuan Fang \\ University of Glasgow \\ \texttt{\small j.fang.2@research.gla.ac.uk} \\ \And
Zaiqiao Meng\thanks{Corresponding Author.} \\ University of Glasgow \\ \texttt{\small zaiqiao.meng@glasgow.ac.uk} \\ \And
Craig Macdonald \\ University of Glasgow \\ \texttt{\small craig.macdonald@glasgow.ac.uk} \\
}

\begin{document}
\maketitle
\begin{abstract}
Retrieval-augmented generation (RAG) offers an effective approach for addressing question answering (QA) tasks. However, the imperfections of the retrievers in RAG models often result in the retrieval of irrelevant information, which could introduce noises and degrade the performance, especially when handling multi-hop questions that require multiple steps of reasoning.  To enhance the multi-hop reasoning ability of RAG models, we propose \textbf{\OURS{}}\footnote{Code: \url{https://github.com/jyfang6/trace}}. \OURS{} constructs \textit{knowledge-grounded reasoning chains}, which are a series of logically connected knowledge triples, to identify and integrate {supporting evidence} from the retrieved documents for answering questions.  Specifically, \OURS{} employs a \textit{KG Generator} to create a knowledge graph (KG) from the retrieved documents, and then uses an \textit{Autoregressive Reasoning Chain Constructor} to build reasoning chains.  Experimental results on three multi-hop QA datasets show that \OURS{} achieves an average performance improvement of up to $14.03\%$ compared to using all the retrieved documents. Moreover, the results indicate that using reasoning chains as context, rather than the entire documents, is {often} sufficient to correctly answer questions. 

\end{abstract}

\section{Introduction}

Retrieval-augmented generation (RAG) models have achieved remarkable performance on question answering (QA) task~\cite{lewis2020retrieval,izacard2023atlas,ram2023context,lin2024ra}. These models employ a \textit{retriever}-\textit{reader} architecture~\cite{karpukhin2020dense}. The \textit{retriever} is used to retrieve a set of documents relevant to the questions, and the \textit{reader} generates answers based on these documents. Moreover, the reader is often instantiated with large language models (LLMs) due to their powerful in-context learning capabilities, leading to superior zero-shot performance. In this setting, the retrieved documents are prepended to the questions, which is used as input to the LLMs to generate answers~\cite{ram2023context}. 

However, simply prepending all the documents returned by the retriever can result in suboptimal performance. This is because existing retrievers are not perfect and often include irrelevant documents in the retrieved set~\cite{yoran2024making}. These irrelevant documents introduce noises, which can mislead the reader and degrade performance~\cite{shi2023large}. This issue is particularly problematic when answering \emph{multi-hop} questions, which involve multiple reasoning steps to obtain the correct answers. Previous study indicates that irrelevant documents can significantly impair the multi-hop reasoning ability of RAG models~\cite{yoran2024making}. 

\begin{figure}[!t]
\begin{center}
\includegraphics[width=0.44\textwidth]{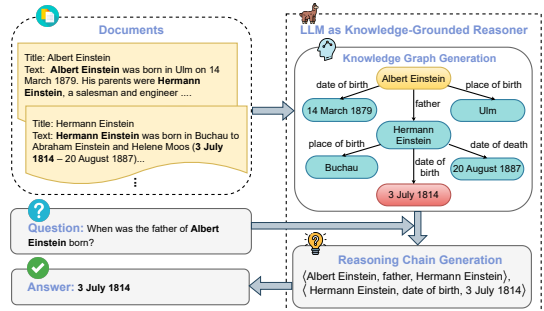}
\end{center}
\vspace{-0.5em}
\caption{\OURS{} transforms documents into a KG and constructs reasoning chains to answer the question.}
\label{figure:flow_chart}
\vspace{-1.0em}
\end{figure}

Therefore, this work focuses on improving the multi-hop reasoning capability of RAG models by enhancing their ability to identify and integrate supporting evidence within documents. The {supporting evidence} refers to the information within documents that directly contributes to answering the questions~\cite{ramesh2023single}. To this end, we propose \textbf{\OURS{}}, which cons\textbf{T}ructs knowledge-grounded \textbf{R}e\textbf{A}soning \textbf{C}hains to identify and integrate supporting \textbf{E}vidence from multiple documents. Figure~\ref{figure:flow_chart} provides an illustration of \OURS{}. 

Specifically, in order to identify supporting evidence, \OURS{} first transforms the retrieved documents into 
a knowledge graph (KG), i.e., a set of knowledge triples in the form of $\langle$\textit{head entity, relation, tail entity}$\rangle$ that describe relationships between entities. This is achieved by using in-context learning to prompt {an LLM instructed as a \textit{KG Generator}} to generate knowledge triples from the retrieved documents. The motivation for converting documents into {a} KG is that, compared with sentences or documents, which contain multiple pieces of information, knowledge triples offer a finer-grained and more concise way to express knowledge, where each triple only conveys a single piece of factual knowledge. Leveraging KG triples can reduce the impact of irrelevant data when identifying {supporting} evidence~\cite{sanmartin2024kg}, leading to more accurate identification of relevant information. 
For example, the sentence ``\textit{Albert Einstein (14 March 1879-18 April 1955) was a German-born theoretical physicist.}'' contains multiple pieces of information about Albert Einstein, including his birth and death days, nationality, and profession. One of the triples generated from this sentence could be ``$\langle$\textit{Albert Einstein, date of birth, 14 March 1879}$\rangle$'', which decouples the birthday information from the sentence. This finer granularity helps in {minimising} the inclusion of irrelevant information, making it easier to identify supporting evidence when answering questions related to Einstein's birthday. 

Notably, the generation of the KG is independent of questions. 
\OURS{} next aims to identify and integrate supporting evidence from the KG to answer multi-hop questions. Specifically, \OURS{} employs an \textit{Autoregressive Reasoning Chain Constructor} to construct reasoning chains from the KG. Each reasoning chain comprises several KG triples that logically connect pieces of supporting evidence to answer the questions. For example, for a multi-hop question ``\textit{When was the father of Albert Einstein born?}'', ``$\langle$\textit{Albert Einstein, father, Hermann Einstein}$\rangle$, $\langle$\textit{Hermann Einstein, date of birth, 3 July 1814}$\rangle$'' is a reasoning chain that provides the necessary information to answer the question. These reasoning chains facilitate the integration of dispersed pieces of supporting evidence, thereby enhancing the model's ability to generate correct answers. 

Moreover, the reasoning chains are constructed in an autoregressive manner. At each step, \OURS{} uses in-context learning to prompt the constructor to select a triple from the KG based on both the question and the previously selected triples. The objective is to ensure that the selected triple forms a logically coherent reasoning chain with the previously selected triples. 
The autoregressive way of constructing reasoning chains is inspired by human reasoning, where each piece of information is considered in the context of what has already been understood. This step-by-step reasoning approach is particularly suitable for multi-hop questions, as it can trace the logical connections across multiple pieces of evidence, ensuring an accurate inference process. For example, in the previously mentioned multi-hop question, if \OURS{} has already identified one piece of supporting evidence: $\langle$\textit{Albert Einstein, father, Hermann Einstein}$\rangle$, it can then focus on finding the next relevant piece of evidence, i.e., $\langle$\textit{Hermann Einstein, date of birth, 3 July 1814}$\rangle$. 

Consequently, compared to vanilla RAG models, \OURS{} creates a KG from the retrieved documents and constructs reasoning chains from the KG in an autoregressive manner to identify and integrate supporting evidence. 
Given the reasoning chains, {the \OURS{} reader} either directly uses them as context to generate the answer (\OURS{}-Triple), or use them to identify a subset of documents that are useful for answering the question (\OURS{}-Doc). 
We conduct experiments on three multi-hop QA datasets in a zero-shot setting and the results show that, compared to using all the retrieved documents, \OURS{}-Triple and \OURS{}-Doc achieve average improvements of 14.03\% and 13.46\% in terms of Exact Match (EM), respectively. 
Moreover, our results indicate that, in the RAG setting, constructing more condensed reasoning chains (i.e., KG triples) from the retrieved documents as context, rather than using the entire documents, is {often} sufficient to correctly answer questions. 

Our contributions are summarised as follows: 
(1) We propose \OURS{}, which builds knowledge-grounded reasoning chains to enhance the multi-hop reasoning ability of RAG models; 
(2) We propose an autoregressive method to construct reasoning chains to identify and integrate {supporting} evidence; 
(3) Experimental results on three multi-hop QA datasets show that \OURS{} achieves average improvement of up to $14.03\%$ in terms of EM compared to using all the retrieved documents. 

\section{Problem Formulation}
This work focuses on tackling multi-hop questions. 
We denote a multi-hop question and its answer as $q$ and $a$, respectively. Each question is associated with a set of $N$ documents: $\mathcal{D}_q = \{d_1, d_2, \dots d_N\}$, which are obtained with a retriever model.  
Following previous work~\cite{trivedi2023interleaving}, the documents are often retrieved from Wikipedia, where each document comprises a title and a text. 
Given the question $q$ and the document set $\mathcal{D}_q$, the goal is to correctly generate the answer $a$. 

\begin{figure*}[!t]
\begin{center}
\includegraphics[width=0.8\textwidth]{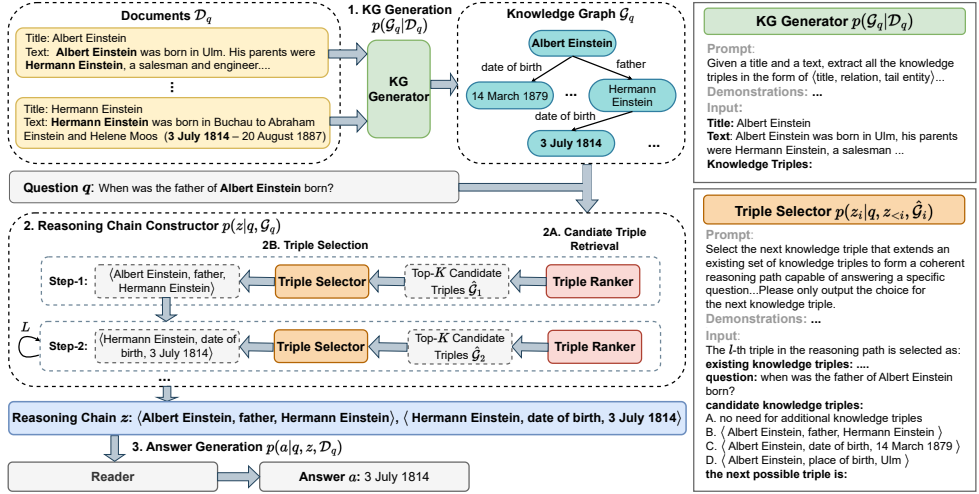}
\end{center}
\vspace{-0.5em}
\caption{Overview of \OURS{}. Given a multi-hop question and the retrieved documents, \OURS{} first uses a KG generator to create a KG based on the documents. It then employs an autoregressive reasoning chain constructor to build reasoning chains from the KG, which are subsequently passed to a reader to generate the answer.}
\label{figure:model}
\vspace{-1.0em}
\end{figure*}

\section{\OURS{}}
This section begins by outlining the overall framework of \OURS{} in \S~\ref{subsec:overall_framework}. Next, we delve into the details of each component in the following sections: KG generator in \S~\ref{subsec:kg_generation}, reasoning chain constructor in \S~\ref{subsec:path_generation}
and finally, answer generation in \S~\ref{subsec:answer_generation}. 

\subsection{Overall Framework}
\label{subsec:overall_framework}
Figure~\ref{figure:model} provides an overview of \OURS{}. 
Given a multi-hop question $q$ and a set of documents $\mathcal{D}_q$, \OURS{} follows these steps to generate the answer: 
\noindent \textbf{(1) KG Generation}: \OURS{} first leverages a \textit{KG generator} to create a KG from $\mathcal{D}_q$, i.e., generating a set of knowledge triples in the form of $\langle$\textit{head entity}, \textit{relation}, \textit{tail entity}$\rangle$;  
\noindent \textbf{(2) Reasoning Chain Construction}: Next, it uses an \textit{autoregressive reasoning chain constructor} to construct reasoning chains \jy{(paths)}\footnote{\jy{We use reasoning chain and reasoning path to denote the same concept, with reasoning path being a commonly used term in the KG reasoning domain~\cite{zhang2022subgraph}.}} from the KG; 
\noindent \textbf{(3) Answer Generation}: \OURS{} generates the answer by either using the reasoning chains directly as context or leveraging the chains to further retrieve their original context documents (see \S~\ref{subsec:answer_generation} for details).

Specifically, \OURS{} can be considered as: 
\begin{align}
    \scalebox{0.88}{$p(a | q, \mathcal{D}_q) \sim p(a|q, z, \mathcal{D}_q) \cdot p(z | q, \mathcal{G}_q) \cdot p(\mathcal{G}_q | \mathcal{D}_q)$},
\end{align}
where $p(\mathcal{G}_q | \mathcal{D}_q)$ denotes the KG generator for creating the KG $\mathcal{G}_q$, $p(z | q, \mathcal{G}_q)$ represents the reasoning chain constructor for building the reasoning chain $z$, which consists of a series of logically connected KG triples, and $p(a|q, z, \mathcal{G}_q)$ denotes the reader that generates the answer. In the following, we introduce the details of each component. 

\subsection{KG Generator}
\label{subsec:kg_generation}

To mitigate the impact of irrelevant data when identifying supporting evidence, \OURS{} employs a {KG generator} $p(\mathcal{G}_q | \mathcal{D}_q)$ to create a KG from $\mathcal{D}_q$. 
Following the recent practice of generating KGs with LLMs~\cite{wei2023zero,zhang2024extract}, we use in-context learning~\cite{wei2022jason} to prompt an LLM instructed as a KG generator to generate KG triples from the documents $\mathcal{D}_q$. 
A straightforward approach is to {concatenate} all the documents within $\mathcal{D}_q$ as inputs and prompt the LLM to generate KG triples. {However, this approach may suffer from the ``lost-in-the-middle'' issue~\cite{liu2024lost}, where the LLM ignores information from documents placed within the long input.} 
Therefore, \OURS{} independently generates KG triples for each document and constructs relationships across documents by common entities shared among these documents. This approach not only mitigates the lost-in-the-middle issue but also allows for offline precomputation of KG triples for all the documents. 

The prompt used by the KG generator to generate KG triples from a document is detailed in Appendix~\ref{app:prompts_for_generating_kg_triples}. In particular, since this work focuses on documents retrieved from Wikipedia\footnote{For documents retrieved from other sources, one can design specific prompts to optimise the KG generation process.}, which consist of a title and a text, we consider the title as an entity and instruct the KG generator to jointly recognise the entities within the text and infer their relationships with the title entity. This approach leverages the natural relevance between the title and the text, as entities within the text are more likely to have meaningful relationships with the title. 
Our empirical findings show that this KG generation approach yields satisfactory performance. 

\subsection{Reasoning Chain Constructor}
\label{subsec:path_generation}

Given the generated KG $\mathcal{G}_q$, \OURS{} next employs a {reasoning chain constructor} $p(z | q, \mathcal{G}_q)$ to identify and integrate supporting evidence by constructing reasoning chains {in an autoregressive manner}: 
\vspace{-0.5em}
\begin{align}
    & p(z | q, \mathcal{G}_q) = \prod_{i=1}^L p(z_i | q, z_{<i}, \hat{\mathcal{G}}_i), \\
    & \hat{\mathcal{G}}_i  = f(q, z_{<i}, \mathcal{G}_q) , \ \forall i = 1, \dots, L ,
\end{align}
where $L$ denotes the maximum length of reasoning chains, $z_i$ denotes the $i$-th triple in the reasoning chains, {$z_{<i}$ represents all the triples preceding the $i$-th triple}, and $\hat{\mathcal{G}}_i$ denotes the candidate triples from which the $i$-th triple is chosen. {Specifically, at each $i$-th step}, the constructor first employs a \textit{triple ranker} $f(q, z_{<i}, \mathcal{G}_q)$ to obtain a set of candidate triples $\hat{\mathcal{G}}_i$ from $\mathcal{G}_q$, and then uses a \textit{triple selector} $p(z_i | q, z_{<i}, \hat{\mathcal{G}}_i)$ to select the $i$-th triple from the candidate set. Therefore, the reasoning chains are constructed by selecting triples one by one from the KG. We next introduce the details of the triple ranker and the triple selector at each $i$-th step. 

\vspace{0.5em} \noindent \textbf{Triple Ranker.} 
At each $i$-th step, triple ranker aims to select a subset of candidate triples from $\mathcal{G}_q$ that are relevant to $q$ and $z_{<i}$. The triple ranker is implemented with a bi-encoder model $\text{Enc}(\cdot)$. Specifically, the triple ranker considers the concatenation of the question $q$ and the previously selected triples $z_{<i}$ as the query. It then independently encodes the query and all the triples within $\mathcal{G}_q$ as:
\begin{align}
    \boldsymbol{q}_i = \text{Enc}(q, z_{<i}), \ \boldsymbol{t}_j = \text{Enc}(t_j), \, \forall t_j \in \mathcal{G}_q,
\end{align}
where $\boldsymbol{q}_i$ denotes the query embedding and $\boldsymbol{t}_j$ is the embedding of a triple $t_j$ within $\mathcal{G}_q$. Subsequently, the triple ranker uses the inner product, i.e., $\boldsymbol{q}_i^\top \boldsymbol{t}_j$, to estimate the relevance of a triple to the query and selects the top-$K$\footnote{The analyses of the effects of $K$ are in Appendix~\ref{app:effects_of_K}.} triples with the highest relevance as the candidate triples for the $i$-th step, i.e., $\hat{\mathcal{G}}_i$.

\vspace{0.5em} \noindent \textbf{Triple Selector.} 
At each $i$-th step, triple selector aims to select a triple from the candidate set $\hat{\mathcal{G}}_i$ to form a coherent reasoning chain with existing triples ($z_{<i}$). We use in-context learning to prompt an LLM instructed as the triple selector to select the triple. 
In particular, we formulate this task as a multiple-choice task, where each candidate triple is formatted as an option, e.g., ``\textit{B.} $\langle$\textit{Albert Einstein, father, Hermann Einstein}$\rangle$''. 
This multiple-choice task formulation can mitigate the hallucination of the LLM, as all the selected triples are grounded in the KG. 
Given the question, the previously selected triples and a list of options, the triple selector is instructed to output a single token representing the option of the selected triple, e.g., ``B''. 
Additionally, we introduce a special ``A'' option for \jy{the chain termination strategy, which will be introduced later.} 
The prompt used by the triple selector to select the $i$-th triple is detailed in Appendix~\ref{app:prompts_for_generating_reasoning_chains}.  

Moreover, at each $i$-th step, we use the logits of the option tokens, which are output by the triple selector, to define a distribution for the $i$-th triple\footnote{We found that calculating the triple distribution at each step and using beam search to generate multiple reasoning chains performs better than selecting one triple at each step, which results in only one reasoning chain (see Appendix~\ref{app:effects_of_R}). }:
\begin{align}
    \scalebox{0.95}{$p(z_i | q, z_{<i}, \hat{\mathcal{G}}_i) = \text{Softmax}(l({c_1}), \dots, l({c_K})),$}
\end{align}
where $c_1$ and $c_K$ represent option tokens and $l(\cdot)$ denotes the logit of the corresponding option token. For example, if there are two candidate triples for the $i$-th step: ``B. $\langle$\textit{Hermann Einstein, date of birth, 3 July 1814}$\rangle$, C. $\langle$\textit{Albert Einstein, born, 14 March 1879}$\rangle$'',
we use the logits of the tokens ``B'' and ``C'' to obtain the distribution for the $i$-th triple. 

\vspace{0.5em} \noindent \textbf{Chain Construction.}
With the triple distributions we use beam search to generate $R$ reasoning chains. At each step of the beam search, we select the top-$b$ triples with the highest probability, as we empirically found that this approach can achieve satisfactory performance. The pseudo code and computational complexity analysis of reasoning chain construction process are in Appendix~\ref{app:algorithm}. 

Moreover, considering that different multi-hop questions may require a varying number of triples in the reasoning chains, we introduce a special option at each step of the reasoning chain generation process: ``A. \textit{no need for additional triples}''. This option indicates that previously selected triples are sufficient to answer the question, and no additional triples are needed. Once the triple selector chooses this option, the generation of the current reasoning chain is terminated. We refer to this approach as the \textit{adaptive chain termination} strategy. 
Our empirical results show that this strategy significantly improves the performance {compared to using fixed-length reasoning chains} (see Appendix~\ref{app:adaptive_chain_termination}).

\subsection{Answer Generation}
\label{subsec:answer_generation}
Finally, \OURS{} leverages a reader to generate the answer, i.e., $p(a|q, z, \mathcal{D}_q)$. 
We propose two methods to use the reasoning chains for generating the answer. 
In the first method, termed \textit{\OURS{}-Triple}, we directly use the reasoning chains as context to generate the answer. 
The second method, termed \textit{\OURS{}-Doc}, uses the triples within the reasoning chains to retrieve their original documents from $\mathcal{D}_q$ and then uses these documents as the context to generate the answer. For both methods, we found that the ordering of the input context is important for the answer generation performance. The constructed reasoning chains naturally present an ideal order of the triples, therefore \textit{\OURS{}-Triple} directly concatenate these triples as the input context, achieving satisfactory performance. However, multiple triples from the reasoning chains could appear in a single document, requiring an alternative ordering method for the retrieved documents in \textit{\OURS{}-Doc}. In practice, we found that a majority voting approach achieves satisfactory results. Specifically, each triple in the reasoning chains casts a vote for the document from which the triple is generated. We aggregate all the votes to rank the documents in $\mathcal{D}_q$ based on the number of votes they receive. Documents that receive no votes are filtered out. 

\section{Experiments}

\subsection{Experimental Setup}
\label{subsec:experimental_setup}

\textbf{Datasets.} 
We conduct experiments using three multi-hop QA datasets: \textbf{HotPotQA}~\cite{yang2018hotpotqa}, \textbf{2WikiMultiHopQA}~\cite{ho2020constructing} and \textbf{MuSiQue}~\cite{trivedi2022musique}. 
These datasets typically require 2-4 hops of reasoning to answer the questions. 
Each question in HotPotQA, 2WikiMultiHopQA and MuSiQue is associated with 10, 10 and 20 documents, respectively, all retrieved from Wikipedia. 
More details and statistics about the datasets are provided in Appendix~\ref{app:datasets}. 

\begin{table*}[tb]
\centering
\resizebox{0.7\textwidth}{!}{
\begin{tabular}{lcccccccccccc}
\toprule
\multirow{2}{*}{\textbf{Model}} & \multicolumn{3}{c}{\textbf{HotPotQA}} & \multicolumn{3}{c}{\textbf{2WikiMultiHopQA}} & \multicolumn{3}{c}{\textbf{MuSiQue}}  \\
\cmidrule(lr){2-4} \cmidrule(lr){5-7} \cmidrule(lr){8-10} 
& \textbf{\# Tok} & \textbf{EM} & \textbf{F1} & \textbf{\# Tok} & \textbf{EM} & \textbf{F1} & \textbf{\# Tok} & \textbf{EM} & \textbf{F1} \\
\midrule\noalign{\vskip 0.1ex}
\multicolumn{10}{l}{\textit{\textbf{LLaMA3 Reader with Context from Naive Baselines}}}\vspace{0.1ex} \\

\textbf{w/o documents} 
& \textcolor{white}{00}56 & 19.28\textcolor{white}{$^*$} &26.81\textcolor{white}{$^*$} 
& \textcolor{white}{00}53 &19.53\textcolor{white}{$^*$} &25.11\textcolor{white}{$^*$} &\textcolor{white}{00}57 & \textcolor{white}{0}3.85\textcolor{white}{$^*$} & \textcolor{white}{0}8.32\textcolor{white}{$^*$} 
\\
\textbf{w. all documents} & 1,430 &45.40\textcolor{white}{$^*$} &60.49\textcolor{white}{$^*$} &1,056 & 28.35\textcolor{white}{$^*$} & 46.07\textcolor{white}{$^*$} &2,551 &16.14\textcolor{white}{$^*$} &23.68\textcolor{white}{$^*$} \\

\cmidrule(lr){1-10}\noalign{\vskip 0.1ex}
\multicolumn{10}{l}{\textit{\textbf{LLaMA3 Reader with Context from Bi-Encoders}}}\vspace{0.1ex} \\
\textbf{Contriever} & 1,430 & 47.10\textcolor{white}{$^*$} & 62.39\textcolor{white}{$^*$} & \textcolor{white}{0}894 & 28.01\textcolor{white}{$^*$} &46.41\textcolor{white}{$^*$} &\textcolor{white}{0}767 &20.23\textcolor{white}{$^*$} &27.86\textcolor{white}{$^*$} \\
\textbf{DRAGON+} & 1,430 & 46.47\textcolor{white}{$^*$} & 61.52\textcolor{white}{$^*$} &\textcolor{white}{0}900 &27.51\textcolor{white}{$^*$} &46.04\textcolor{white}{$^*$} &1,387 &19.94\textcolor{white}{$^*$} & 27.21\textcolor{white}{$^*$} \\
\textbf{E5} & \textcolor{white}{0}920 &47.52\textcolor{white}{$^*$} &62.96\textcolor{white}{$^*$} &\textcolor{white}{0}633 &29.02\textcolor{white}{$^*$} &46.52\textcolor{white}{$^*$} & 1,079 & 24.08\textcolor{white}{$^*$} & 31.71\textcolor{white}{$^*$} \\
\textbf{E5-Mistral} & 1,430 &48.25\textcolor{white}{$^*$} &63.72\textcolor{white}{$^*$} &\textcolor{white}{0}639 &31.44\textcolor{white}{$^*$} &48.83\textcolor{white}{$^*$} &1,099 &26.40\textcolor{white}{$^*$} & 33.66\textcolor{white}{$^*$} \\

\cmidrule(lr){1-10}\noalign{\vskip 0.1ex}
\multicolumn{10}{l}{\textit{\textbf{LLaMA3 Reader with Context from Cross-Encoders}}}\vspace{0.1ex} \\
\textbf{monoT5} & 1,430 &47.10\textcolor{white}{$^*$} &62.39\textcolor{white}{$^*$} &\textcolor{white}{0}654 &29.84\textcolor{white}{$^*$} &47.46\textcolor{white}{$^*$} & 1,121 &24.12\textcolor{white}{$^*$} &31.96\textcolor{white}{$^*$} \\
\textbf{BGE} & 1,430 &48.40\textcolor{white}{$^*$} &63.65\textcolor{white}{$^*$} &\textcolor{white}{0}481 &32.64\textcolor{white}{$^*$} &48.93\textcolor{white}{$^*$} & 1,406 &25.53\textcolor{white}{$^*$} &33.27\textcolor{white}{$^*$} \\
\textbf{RankLLaMA} & 1,430 & 46.41\textcolor{white}{$^*$} & 61.74\textcolor{white}{$^*$} &\textcolor{white}{0}474 &32.46\textcolor{white}{$^*$} &48.19\textcolor{white}{$^*$} &1,189 & 23.00\textcolor{white}{$^*$} &29.89\textcolor{white}{$^*$} \\

\cmidrule(lr){1-10}\noalign{\vskip 0.1ex}
\multicolumn{10}{l}{\textit{\textbf{LLaMA3 Reader with Context from Chain-of-Thought Model}}}\vspace{0.1ex} \\
\textbf{IRCoT} &\textcolor{white}{0}454	&50.78\textcolor{white}{$^*$} &65.65\textcolor{white}{$^*$} &\textcolor{white}{0}553 &36.11\textcolor{white}{$^*$} & 52.25\textcolor{white}{$^*$} &\textcolor{white}{0}571 & 27.40\textcolor{white}{$^*$} &36.91\textcolor{white}{$^*$} \\ 

\cmidrule(lr){1-10}\noalign{\vskip 0.1ex}
\multicolumn{10}{l}{\textit{\textbf{LLaMA3 Reader with Context from \OURS{}}}}\vspace{0.1ex} \\
\textbf{\OURS{}-Triple} &\textcolor{white}{0}160 & 53.05$^*$ & 67.32$^*$ &\textcolor{white}{0}164 &\textbf{45.51}$^*$ &\textbf{56.43}$^*$ &\textcolor{white}{0}169 & \textbf{33.43}$^*$ &\textbf{40.05}$^*$ \\ 
\textbf{\OURS{}-Doc} &\textcolor{white}{0}357 & \textbf{55.08}$^*$ & \textbf{69.99}$^*$ &\textcolor{white}{0}485 & 42.74$^*$ &{55.30}$^*$ &\textcolor{white}{0}456 &32.44$^*$ & 40.03$^*$ \\  
\bottomrule    
\end{tabular}
} 
\vspace{-0.5em}
\caption{Overall performance (\%) of \OURS{} and baselines on the test sets of multi-hop QA datasets, where ``\# Tok'' is the average number of tokens in the documents/reasoning chains used as context, $^*$ indicates p-value < 0.05 compared with IRCoT. The best performance per dataset per metric is marked in boldface.}
\label{table:overall_performance}
\vspace{-0.5em}
\end{table*}

\vspace{0.5em} \noindent \textbf{Baselines.}
Since \OURS{} builds reasoning chains from $\mathcal{D}_q$ to enhance RAG models, we mainly compare it against naive RAG baselines as well as other baselines capable of recognising supporting documents within $\mathcal{D}_q$: 
(1) naive baselines: \textit{w/o documents}, where no documents are used in reader models, \textit{w. all documents}, where all the documents are used in readers (i.e., the vanilla RAG);
(2) bi-encoders: {Contriever}~\cite{gautier2022unsupervised}, {E5}~\cite{wang2022text}, {DRAGON+}~\cite{lin2023how}, {E5-Mistral}~\cite{wang2023improving}; 
(3) cross-encoders: {monoT5}~\cite{nogueira2020document}, {RankLLaMA}~\cite{ma2023fine}, {BGE}~\cite{xiao2023c}; 
(4) chain-of-thought (CoT): {IRCoT}~\cite{trivedi2023interleaving}. 
Particularly, for bi-encoders and cross-encoders, we use them to rank the documents in $\mathcal{D}_q$ based on their estimated relevance scores. The top-$M$ documents are used as context, where $M$ is selected from $\{1, \dots, N\}$ and set to the number of documents that results in the best performance on the development set of each dataset. 
For fair comparisons, both \OURS{} and baselines use the same reader to generate answers. 
More details about the baselines are provided in Appendix~\ref{app:baselines}. 

\vspace{0.5em} \noindent \textbf{Evaluation.} 
To evaluate the QA performance, we follow previous work~\cite{ramesh2023single} and use Exact Match (EM) and F1 as evaluation metrics, which are the standard metrics for these datasets. 

\vspace{0.5em} \noindent \textbf{Implementation Details.} 
{\tt LLaMA3-8B-Instruct} is used for both the KG generator and the triple selector. 
We employ {\tt e5-mistral-7b-instruct} as the triple ranker. 
The analyses of the generated KGs and reasoning chains are in Appendix~\ref{app:statistics_kg_chains}. 
For the reader, we use in-context learning to generate answers in a zero-shot setting (see Appendix~\ref{app:prompts_for_generating_answers}), 
and report the performance of different readers, such as {\tt LLaMA3-8B-Instruct}, {\tt Mistral-7B-v0.1} and {\tt Gemma-7B}. We mainly report the performance using LLaMA3 as the reader unless otherwise stated. Details about prompts and hyperparameters can be found in Appendix~\ref{app:implementation}. 

\subsection{Results and Analysis}
We provide our main experimental results in this section. Additional results are in Appendix~\ref{app:additional_results_analysis}.

%%%%%%%%%%%%%%%%%%%%%%%%%%%%%%%%%
% Research Question 1 
%%%%%%%%%%%%%%%%%%%%%%%%%%%%%%%%%
\vspace{0.5em} \noindent \textbf{(RQ1): How does \OURS{} perform against the baselines?} 
The QA performance of \OURS{} and baselines is reported in Table~\ref{table:overall_performance}, yielding the following findings: 
(1) First, \OURS{}-Triple and \OURS{}-Doc consistently achieve the best performance on all the datasets, demonstrating the effectiveness of \OURS{} in multi-hop QA tasks. Compared to the vanilla RAG model (i.e., \textit{w. all documents}) \OURS{}-Triple and \OURS{}-Doc achieve average improvements of $14.03\%$ and $13.46\%$ in terms of EM across all datasets, respectively.  
This superior performance is due to \OURS{}'s ability to effectively identify supporting evidence within the documents while avoiding the introduction of irrelevant context; 
(2) Moreover, compared to the CoT model IRCoT, the best performing baseline, \OURS{}-Triple and \OURS{}-Doc achieve significantly better performance, with average improvements of $5.90\%$ and $5.32\%$ in EM, respectively. The suboptimal performance of IRCoT may be due to LLMs' tendency to hallucinate and generate inaccurate CoT sentences~\cite{luo2023reasoning}. 
In contrast, the KG triples in \OURS{}'s reasoning chains are grounded in the documents and selected for their relevance to the questions, ensuring a more reliable and accurate reasoning process; 
(3) Furthermore, \OURS{}-Triple, which only uses reasoning chains as context, achieves the best performance on the 2WikiMultiHopQA and MuSiQue datasets, and the second-best performance on the HotPotQA dataset in EM. 
Notably, the reasoning chains used by \OURS{}-Triple contain fewer tokens compared to the documents used by other models, yet it achieves the best performance in most cases. 
This result indicates that it is unnecessary to use all the information within documents; instead, constructing reasoning chains to identify and integrate supporting evidence within the documents is often sufficient to correctly answer questions. 

Additionally, we report the QA performance of \OURS{} using different \jy{reader models}. Due to the space limit, we provide the EM performance of vanilla RAG (\textit{w. all documents}) and our models on the HotPotQA and 2WikiMultiHopQA datasets in Figure~\ref{figure:generalisation_on_different_readers}. The complete results for different readers are in Appendix~\ref{app:overall_performance_different_readers}. The results in Figure~\ref{figure:generalisation_on_different_readers} indicate that both \OURS{}-Triple and \OURS{}-Doc outperform the baseline by a large margin when using different readers, demonstrating that the reasoning chains generated by \OURS{} can effectively generalise across various readers. 

\begin{figure}[tb]
\begin{center}
\includegraphics[width=0.45\textwidth]{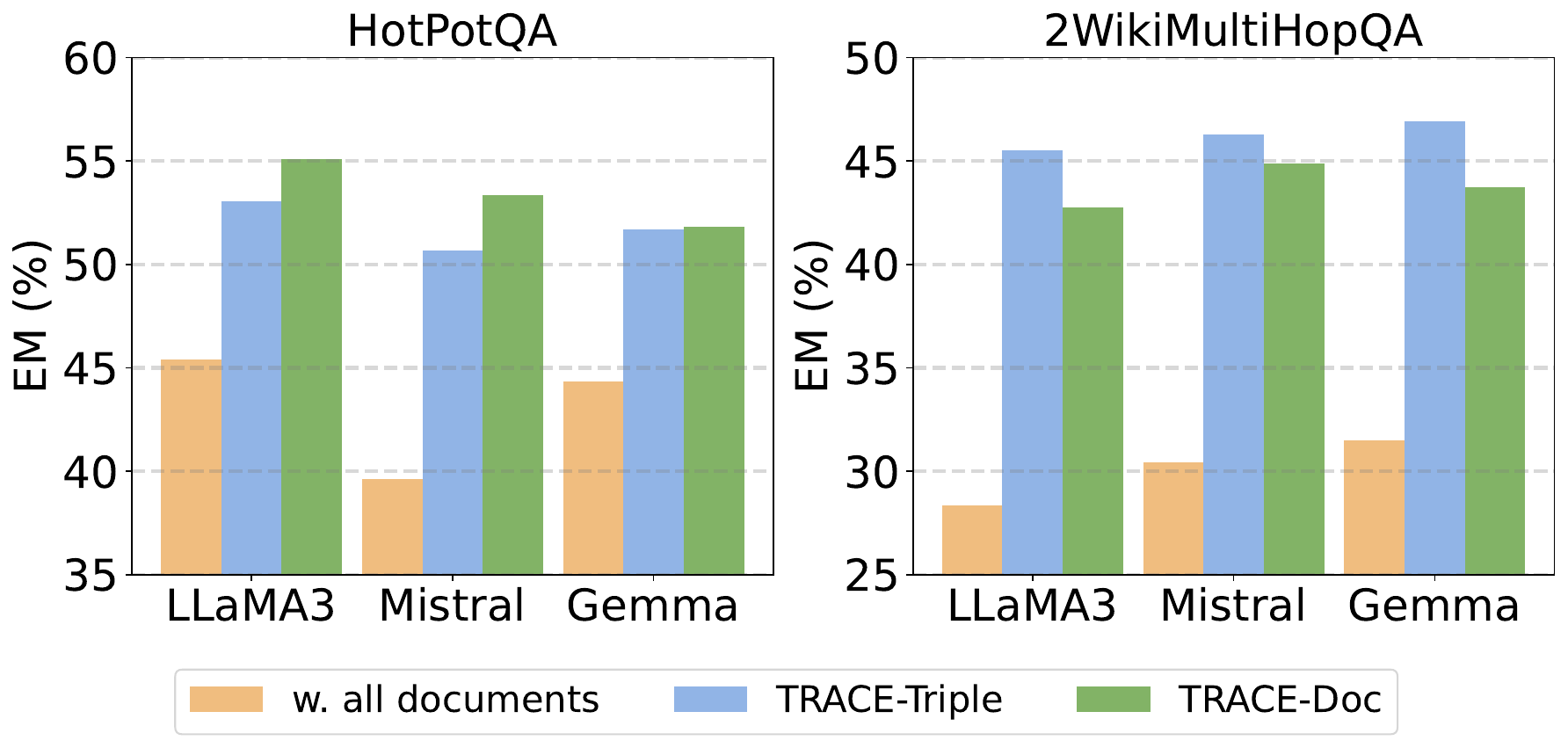}
\end{center}
\vspace{-0.5em}
\caption{Performance of \OURS{} with different readers on the test sets of HotPotQA and 2WikiMultiHopQA.}
\label{figure:generalisation_on_different_readers}
\vspace{-1.0em}
\end{figure}

\begin{table}[tb]
\centering
\resizebox{0.48\textwidth}{!}{
\begin{tabular}{lcccccc}
\toprule
\multirow{2}{*}{\textbf{Model}} & \multicolumn{2}{c}{\textbf{HotPotQA}} & \multicolumn{2}{c}{\textbf{2WikiMultiHopQA}} & \multicolumn{2}{c}{\textbf{MuSiQue}}  \\
\cmidrule(lr){2-3} \cmidrule(lr){4-5} \cmidrule(lr){6-7} 
& \textbf{EM} & \textbf{F1} & \textcolor{white}{00}\textbf{EM} & \textbf{F1} & \textbf{EM} & \textbf{F1}  \\
\midrule
\textbf{\OURS{}-Triple} & \textbf{67.00}\textcolor{white}{$^*$} & \textbf{76.16}\textcolor{white}{$^*$} & \textcolor{white}{00}\textbf{66.40}\textcolor{white}{$^*$} & \textbf{72.52}\textcolor{white}{$^*$} & \textbf{40.40}\textcolor{white}{$^*$} & \textbf{46.92}\textcolor{white}{$^*$} \\

\cmidrule(lr){1-7}\noalign{\vskip 0.5ex}
\multicolumn{7}{l}{\textit{\textbf{Effectiveness of KG Generator}}}\vspace{0.3em} \\ 
\textbf{w. sentences} & 65.60$^*$ & 74.72$^*$ & \textcolor{white}{00}63.40$^*$ & 71.00$^*$ & 35.40$^*$ & 41.95$^*$ \\
\textbf{w. documents} & 65.00$^*$ & 75.93$^*$ & \textcolor{white}{00}58.60$^*$ & 68.87$^*$ & 30.80$^*$ & 35.84$^*$ \\

\cmidrule(lr){1-7}\noalign{\vskip 0.5ex}
\multicolumn{7}{l}{\textit{\textbf{Effectiveness of Reasoning Chain Constructor}}}\vspace{0.3em} \\ 
\textbf{w. Top-10 Triples} & 55.40$^*$ & 65.67$^*$ & \textcolor{white}{00}43.80$^*$ & 47.99$^*$ & 30.00$^*$ & 36.44$^*$ \\
\textbf{w. Top-20 Triples} & 59.40$^*$ & 69.84$^*$ & \textcolor{white}{00}48.00$^*$ & 53.12$^*$ & 30.00$^*$ & 36.77$^*$ \\
\textbf{w. Top-25 Triples} & 59.40$^*$ & 70.29$^*$ & \textcolor{white}{00}48.40$^*$ & 54.13$^*$ & 30.60$^*$ & 37.03$^*$ \\

\cmidrule(lr){1-7}\noalign{\vskip 0.5ex}
\multicolumn{7}{l}{\textit{\textbf{Effectiveness of Triple Ranker}}}\vspace{0.3em} \\ 
\textbf{w/o Triple Ranker} & 58.40$^*$ & 67.71$^*$ & \textcolor{white}{00}63.40$^*$ & 69.05$^*$ & 26.40$^*$ & 34.55$^*$ \\ 
\textbf{w. DRAGON+} & 62.80$^*$ & 72.86$^*$ & \textcolor{white}{00}64.40$^*$ & 70.55$^*$ & 36.30$^*$ & 42.16$^*$ \\
\textbf{w. E5} & 64.00$^*$ & 73.88$^*$ & \textcolor{white}{00}65.20\textcolor{white}{$^*$} & 71.10\textcolor{white}{$^*$} & 36.60$^*$ & 43.18$^*$ \\

\cmidrule(lr){1-7}\noalign{\vskip 0.5ex}
\multicolumn{7}{l}{\textit{\textbf{Effectiveness of Triple Selector}}}\vspace{0.3em} \\ 
\textbf{w/o Triple Selector} & 53.80$^*$ & 63.80$^*$ & \textcolor{white}{00}47.00$^*$ & 52.31$^*$ & 26.20$^*$ & 31.24$^*$ \\ 
\textbf{w. Mistral} & 61.80$^*$ & 71.04$^*$ & \textcolor{white}{00}65.60\textcolor{white}{$^*$} & 72.01\textcolor{white}{$^*$} & 34.60$^*$ & 41.49$^*$ \\ 
\textbf{w. Gemma} &59.00$^*$ & 68.52$^*$ & \textcolor{white}{00}63.40$^*$ & 69.26$^*$ & 28.80$^*$ & 35.08$^*$ \\ 

\bottomrule    
\end{tabular}
} 
\vspace{-0.5em}
\caption{Performance (\%) of \OURS{}-Triple and its variants on the development sets of three QA datasets, where $^*$ denotes p<0.05 compared with \OURS{}-Triple.}
\label{table:ablation_study_use_triple}
\vspace{-1.0em}
\end{table}

%%%%%%%%%%%%%%%%%%%%%%%%%%%%%%%%%
% Research Question 2 
%%%%%%%%%%%%%%%%%%%%%%%%%%%%%%%%%
\vspace{0.5em} \noindent \textbf{(RQ2): What is the advantage of generating KGs from documents?} 
To investigate the advantage of generating KGs, we introduce two variants of \OURS{}-Triple: \textit{w. sentences} and \textit{w. documents}, where the KG generator is removed and the triples are replaced with sentences and documents from $\mathcal{D}_q$, respectively. These sentences or documents are passed to the reasoning chain constructor to build sentence-based or document-based reasoning chains. The QA performance of \OURS{}-Triple and its variants is reported in Table~\ref{table:ablation_study_use_triple}. The results show that \OURS{}-Triple significantly outperforms the two variants on all datasets. This is because, compared with sentences or documents, KG triples provide a finer-grained and more concise way of storing knowledge, containing less irrelevant information. Therefore, using KG triples to identify supporting evidence can mitigate the impact of irrelevant information, improving the accuracy of the reasoning process. 
We conduct the same experiments on \OURS{}-Doc. The results are in Appendix~\ref{app:ablation_of_trace_doc}, which demonstrate similar outcomes.

%%%%%%%%%%%%%%%%%%%%%%%%%%%%%%%%%
% Research Question 3 
%%%%%%%%%%%%%%%%%%%%%%%%%%%%%%%%%
\vspace{0.5em} \noindent \textbf{(RQ3): Can the reasoning chain constructor effectively identify supporting evidence?}
To investigate the effectiveness of the chain constructor, we introduce a variant of \OURS{}-Triple: \textit{w. Top-$T$ Triples}, where we remove the constructor and directly use E5-Mistral to retrieve the top-$T$ most relevant triples from the KG. These triples are used in a manner similar to reasoning chains.  
The results of \OURS{}-Triple and the variants are reported in Table~\ref{table:ablation_study_use_triple}, which indicate that removing the chain constructor significantly degrades the performance on all the datasets. This is because the reasoning chain constructor identifies supporting evidence in an autoregressive manner. The previously identified supporting evidence  provides cues for identifying the subsequent evidence, thereby facilitating the identification of all the supporting evidence and leading to improved performance.

\begin{figure}[tb]
\begin{center}
\includegraphics[width=0.45\textwidth]{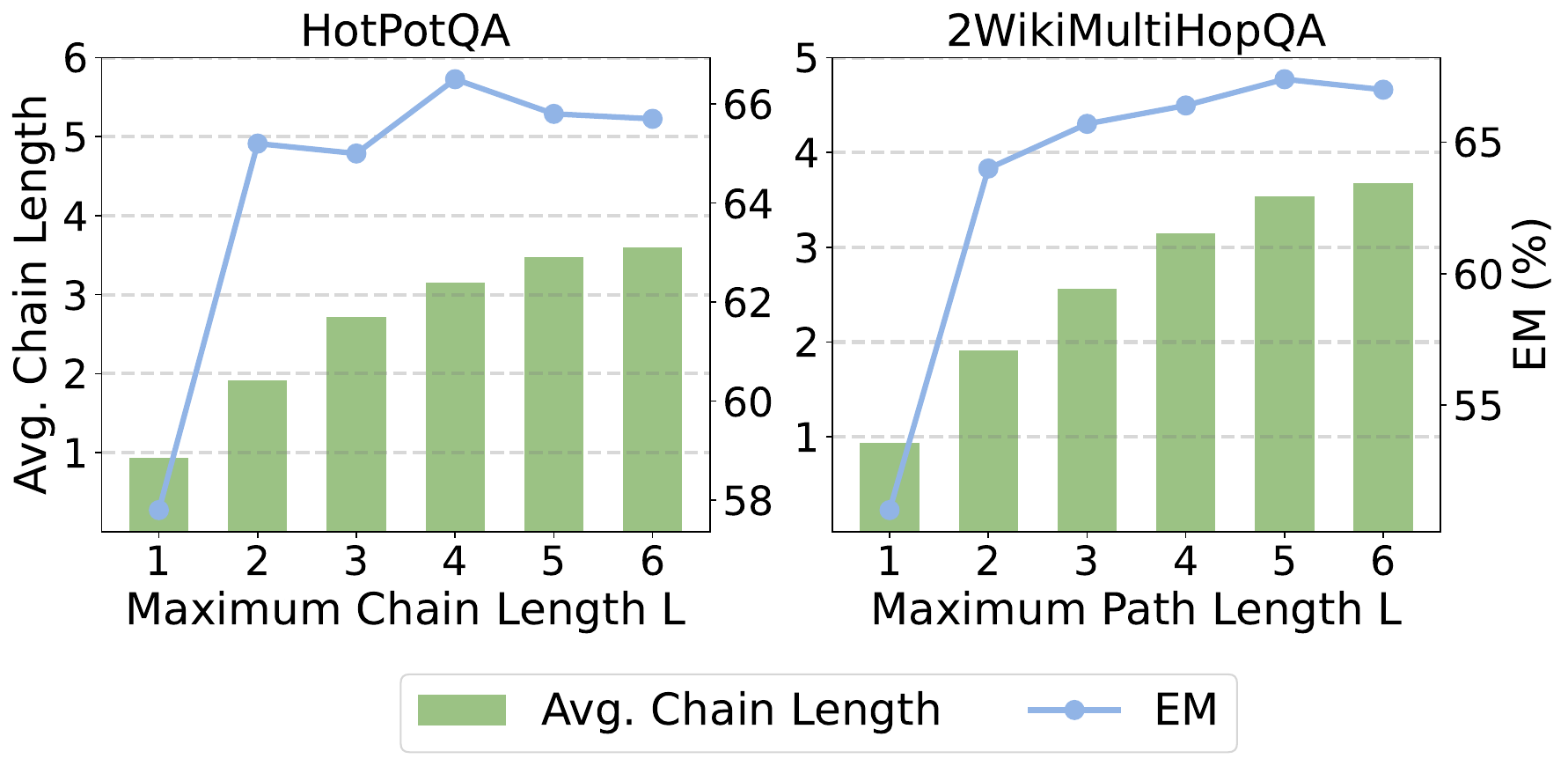}
\end{center}
\vspace{-0.5em}
\caption{QA performance~(\%) and average chain length of \OURS{}-Triple under different values of $L$ on the development sets of HotPotQA and 2WikiMultiHopQA.}
\label{figure:maximum_path_length_part12}
\vspace{-1.0em}
\end{figure}

%%%%%%%%%%%%%%%%%%%%%%%%%%%%%%%%%
% Research Question 4
%%%%%%%%%%%%%%%%%%%%%%%%%%%%%%%%%
\vspace{0.5em} \noindent \textbf{(RQ4): What are the effects of each component, i.e., triple ranker and triple selector, in the reasoning chain constructor?}
To examine the effectiveness of the triple ranker, we introduce a variant of \OURS{}-Triple: \textit{w/o Triple Ranker}, where the triple ranker is removed and $\mathcal{G}_q$ is used as the candidate set. 
The results in Table~\ref{table:ablation_study_use_triple} indicate that removing the triple ranker significantly degrades the performance on all the datasets. The superior performance of using the triple ranker is due to its ability to identify triples that are relevant to the current context while avoiding the introduction of irrelevant ones. 
We next examine the impact of different choices for the triple ranker. In addition to the E5-Mistral model used in \OURS{}-Triple, we also test DRAGON+ and E5 as the triple ranker, which we empirically found to be more effective than other models, denoted as \textit{w. DRAGON+} and \textit{w. E5}, respectively. The results in Table~\ref{table:ablation_study_use_triple} show that different triple rankers impact the final performance, with E5-Mistral achieving the best results. 

Moreover, to investigate the effectiveness of the triple selector, we introduce a variant of \OURS{}-Triple: \textit{w/o Triple Selector}, where the triple selector is removed and the top-$b$ triples ranked by the triple ranker are used to construct reasoning chains.  
The results in Table~\ref{table:ablation_study_use_triple} show that removing the triple selector significantly degrades performance on all the datasets, indicating the importance of using the triple selector's reasoning ability in constructing coherent reasoning chains. 
We also examine the impact of different choices for the triple selector. 
In addition to LLaMA3 used in \OURS{}-Triple, we also use Mistral and Gemma as the triple selector, denoted as \textit{w. Mistral} and \textit{w. Gemma}, respectively. 
The results in Table~\ref{table:ablation_study_use_triple} suggest that different triple selectors also affect the final performance, with LLaMA3 achieving the best performance.

\begin{table}[tb]
\centering
\begin{small}
\begin{tabular}{p{0.45\textwidth}}
\toprule
\textbf{Question}: Are both Blaise Cendrars and Julian Barnes are a citizen of the same country? \\
\textbf{Reasoning Chain}: $\langle$Blaise Cendrars; nationality; Swiss$\rangle$, $\langle$Julian Barnes; nationality; English$\rangle$ \\
\textbf{Generated Answer}: no \\

\noalign{\vskip 0.5ex}\hdashline\noalign{\vskip 0.5ex}

\textbf{Question}: What was the occupation of both Christina Stead and Nuruddin Farah? \\
\textbf{Reasoning Chain}: $\langle$Christina Stead; occupation; novelist and short-story writer$\rangle$, $\langle$Nuruddin Farah; occupation; novelis$\rangle$ \\
\textbf{Generated Answer}: novelist \\

\noalign{\vskip 0.5ex}\hdashline\noalign{\vskip 0.5ex}

\textbf{Question}: What is the birth date of this Spanish footballer, who was added as a holding midfielder in the 2012-13 FC
Bayern Munich season? \\
\textbf{Reasoning Chain}: $\langle$2012–13 FC Bayern Munich season; new player signed after the first week of the Bundesliga season; Javi Martínez$\rangle$,
$\langle$Javi Martínez; date of birth; 2 September 1988$\rangle$ \\
\textbf{Generated Answer}: 2 September 1988 \\

\bottomrule    
\end{tabular}
\end{small}
\vspace{-0.5em}
\caption{Case Study of \OURS{} on HotPotQA dataset.}
\label{table:case_study}
\vspace{-1.0em}
\end{table}

%%%%%%%%%%%%%%%%%%%%%%%%%%%%%%%%%
% Research Question 5 
%%%%%%%%%%%%%%%%%%%%%%%%%%%%%%%%%
\vspace{0.5em} \noindent \textbf{(RQ5): How does the maximum chain length $L$ affect the performance?} 
We conduct experiments to investigate the effects of the hyperparameter $L$. Specifically, we vary the value of $L$ from 1 to 6 and report the corresponding results, including the average chain length and the QA performance. Figure~\ref{figure:maximum_path_length_part12} shows the results of \OURS{}-Triple on HotPotQA and 2WikiMultiHopQA. The results on MuSiQue are in Appendix~\ref{app:effects_of_l_on_musique}. The figure shows that when increasing $L$ from 1 to 6, the average chain length does not increase linearly; instead, the growth of the average chain length gradually slows down. This is due to the effectiveness of our adaptive chain termination strategy in automatically determining the optimal lengths of reasoning chains. Moreover, the figure also shows that as we increase $L$, the QA performance initially increases but then decreases after a certain threshold, such as $4$ on HotPotQA. This decline in performance may be due to the introduction of irrelevant or redundant information in longer reasoning chains, which can confuse the reader and degrade the performance.

%%%%%%%%%%%%%%%%%%%%%%%%%%%%%%%%%
% Case Study 
%%%%%%%%%%%%%%%%%%%%%%%%%%%%%%%%%

\vspace{0.5em} \noindent \textbf{Case Study.} 
We conduct a case study to investigate the reasoning chains generated by \OURS{}. 
Table~\ref{table:case_study} shows the reasoning chains and final answers for questions on HotPotQA\footnote{The complete reasoning chains and the identified relevant documents \jy{for these questions} can be found in Appendix~\ref{app:case_study}.}. These examples demonstrate that \OURS{} can construct effective reasoning chains to answer multi-hop questions. Moreover, these reasoning chains also provide an interpretable framework for understanding how the model generates its answers, enhancing the transparency and reliability of the reasoning process.

\section{Related Work}

\textbf{RAG Models.} 
RAG models have demonstrated impressive performance in QA tasks~\cite{karpukhin2020dense,lewis2020retrieval,izacard2020leveraging,ram2023context,shi2023replug}. Recently, RAG models have been used to address multi-hop questions~\cite{yavuz2022modeling,ramesh2023single}. For instance, the IRCoT~\cite{trivedi2023interleaving} uses CoT sentences to retrieve documents and generate answers. Since the LLMs are prone to hallucinate, the generated CoT sentences may be inaccurate~\cite{luo2023reasoning,nguyen2024direct} and lead to suboptimal performance. In contrast, our reasoning chains are grounded on the retrieved documents, ensuring a more reliable and accurate reasoning process. 
Moreover, recent works have shown that the retrieval of irrelevant documents can hinder the performance of RAG models~\cite{petroni2020how,creswell2023selection,shi2023large}, especially when tackling multi-hop questions~\cite{yoran2024making}. Previous works require training to mitigate the effects of irrelevant documents~\cite{ramesh2023single,yoran2024making}, while our \OURS{} uses in-context learning and does not require training. 

\vspace{0.5em} \noindent \textbf{RAG Models with KGs.} 
RAG models have been used to address the knowledge graph question answering task~\cite{jiang2023unikgqa,luo2023reasoning}, where the answers are restricted to be entities in an existing KG, while our work focuses on a more general setting where the answers can be any words or phrases. Under this setting, there are some works that leverage KGs to enhance RAG models~\cite{min2019knowledge,zhou2020knowledge,oguz2022unik,yu2022kg,ju2022grape}. 
For example, UniK-QA~\cite{oguz2022unik} combine KGs into a corpus for retrieval. 
However, these works all use information from existing KGs, such as Wikidata~\cite{vrandevcic2014wikidata}, while \OURS{} creates a KG based on the retrieved documents and constructs reasoning chains for multi-hop reasoning.

\section{Conclusion}
This paper proposes \OURS{} to enhance the multi-hop reasoning ability of RAG models. 
Specifically, \OURS{} employs a KG generator to create a KG from the retrieved documents and then uses an autoregressive reasoning chain constructor to build reasoning chains from the KG. 
Given the reasoning chains, \OURS{} either directly uses them as context to generate answers or uses them to identify a subset of relevant documents. 
\jy{Experimental results on three multi-hop QA datasets show that, compared to using all the retrieved documents, \OURS{} achieves an average performance of up to $14.03\%$ in EM. Moreover, the reasoning chains can effectively generalise across various reader models.} 

\clearpage
\section*{Limitations}
We identify the following limitations of our work: 
(1) Since this work focuses on documents retrieved from Wikipedia, the KG generator primarily generates KG triples between the title and entities within the texts, rather than between all possible entities. 
This approach simplifies the task, which would otherwise be \jy{more} challenging, and we defer such exploration to future work; 
(2) Due to the lack of annotated data, we are unable to directly and quantitatively evaluate the quality the generated KG triples and the constructed reasoning chains.  
Instead, we assess their performance through the final QA performance. 
\jy{Moreover, we provide the statistics of the generated KGs and reasoning chains, as well as some qualitative results in Appendix~\ref{app:statistics_kg_chains} to offer some insights into their structure and effectiveness.}  
We consider improving the evaluation methods for these two modules as one of our future work directions; 
(3) In the reasoning chain constructor, \OURS{} requires access to the logits of the triple selector to define a triple distribution. However, this may not be feasible when using a black-box LLM as the triple selector. In such cases, \OURS{} can use the option token output by the triple selector to select one triple at each step, leading to a single reasoning chain. 

\bibliography{custom}

\begin{thebibliography}{41}
\expandafter\ifx\csname natexlab\endcsname\relax\def\natexlab#1{#1}\fi

\bibitem[{Creswell et~al.(2023)Creswell, Shanahan, and
  Higgins}]{creswell2023selection}
Antonia Creswell, Murray Shanahan, and Irina Higgins. 2023.
\newblock Selection-inference: Exploiting large language models for
  interpretable logical reasoning.
\newblock In \emph{The Eleventh International Conference on Learning
  Representations}.

\bibitem[{Gautier et~al.(2022)Gautier, Mathilde, Lucas, Sebastian, Piotr,
  Armand, and Edouard}]{gautier2022unsupervised}
Izacard Gautier, Caron Mathilde, Hosseini Lucas, Riedel Sebastian, Bojanowski
  Piotr, Joulin Armand, and Grave Edouard. 2022.
\newblock Unsupervised dense information retrieval with contrastive learning.
\newblock \emph{Transactions on Machine Learning Research}.

\bibitem[{Ho et~al.(2020)Ho, Nguyen, Sugawara, and Aizawa}]{ho2020constructing}
Xanh Ho, Anh{-}Khoa~Duong Nguyen, Saku Sugawara, and Akiko Aizawa. 2020.
\newblock Constructing {A} multi-hop {QA} dataset for comprehensive evaluation
  of reasoning steps.
\newblock In \emph{Proceedings of the 28th International Conference on
  Computational Linguistics}, pages 6609--6625.

\bibitem[{Izacard and Grave(2021)}]{izacard2020leveraging}
Gautier Izacard and Edouard Grave. 2021.
\newblock Leveraging passage retrieval with generative models for open domain
  question answering.
\newblock In \emph{Proceedings of the 16th Conference of the European Chapter
  of the Association for Computational Linguistics}, pages 874--880.

\bibitem[{Izacard et~al.(2023)Izacard, Lewis, Lomeli, Hosseini, Petroni,
  Schick, Dwivedi{-}Yu, Joulin, Riedel, and Grave}]{izacard2023atlas}
Gautier Izacard, Patrick S.~H. Lewis, Maria Lomeli, Lucas Hosseini, Fabio
  Petroni, Timo Schick, Jane Dwivedi{-}Yu, Armand Joulin, Sebastian Riedel, and
  Edouard Grave. 2023.
\newblock Atlas: Few-shot learning with retrieval augmented language models.
\newblock \emph{Journal of Machine Learning Research}, 24:251:1--251:43.

\bibitem[{Jiang et~al.(2023)Jiang, Zhou, Zhao, and Wen}]{jiang2023unikgqa}
Jinhao Jiang, Kun Zhou, Xin Zhao, and Ji{-}Rong Wen. 2023.
\newblock $\text{UniKGQA}$: Unified retrieval and reasoning for solving
  multi-hop question answering over knowledge graph.
\newblock In \emph{The Eleventh International Conference on Learning
  Representations}.

\bibitem[{Ju et~al.(2022)Ju, Yu, Zhao, Zhang, and Ye}]{ju2022grape}
Mingxuan Ju, Wenhao Yu, Tong Zhao, Chuxu Zhang, and Yanfang Ye. 2022.
\newblock $\text{GRAPE}$: Knowledge graph enhanced passage reader for
  open-domain question answering.
\newblock In \emph{Findings of the Association for Computational Linguistics},
  pages 169--181.

\bibitem[{Karpukhin et~al.(2020)Karpukhin, Oguz, Min, Lewis, Wu, Edunov, Chen,
  and Yih}]{karpukhin2020dense}
Vladimir Karpukhin, Barlas Oguz, Sewon Min, Patrick S.~H. Lewis, Ledell Wu,
  Sergey Edunov, Danqi Chen, and Wen{-}tau Yih. 2020.
\newblock Dense passage retrieval for open-domain question answering.
\newblock In \emph{Proceedings of the 2020 Conference on Empirical Methods in
  Natural Language Processing}, pages 6769--6781.

\bibitem[{Lewis et~al.(2020)Lewis, Perez, Piktus, Petroni, Karpukhin, Goyal,
  K{\"u}ttler, Lewis, Yih, Rockt{\"a}schel et~al.}]{lewis2020retrieval}
Patrick Lewis, Ethan Perez, Aleksandra Piktus, Fabio Petroni, Vladimir
  Karpukhin, Naman Goyal, Heinrich K{\"u}ttler, Mike Lewis, Wen-tau Yih, Tim
  Rockt{\"a}schel, et~al. 2020.
\newblock Retrieval-augmented generation for knowledge-intensive $\text{NLP}$
  tasks.
\newblock \emph{Advances in Neural Information Processing Systems},
  33:9459--9474.

\bibitem[{Li et~al.(2023)Li, Lv, Yan, Lin, Zhu, Ni, Xie, Wang, and
  Qiu}]{li2023unified}
Xiaonan Li, Kai Lv, Hang Yan, Tianyang Lin, Wei Zhu, Yuan Ni, Guotong Xie,
  Xiaoling Wang, and Xipeng Qiu. 2023.
\newblock Unified demonstration retriever for in-context learning.
\newblock \emph{arXiv preprint arXiv:2305.04320}.

\bibitem[{Lin et~al.(2023)Lin, Asai, Li, Oguz, Lin, Mehdad, Yih, and
  Chen}]{lin2023how}
Sheng{-}Chieh Lin, Akari Asai, Minghan Li, Barlas Oguz, Jimmy Lin, Yashar
  Mehdad, Wen{-}tau Yih, and Xilun Chen. 2023.
\newblock How to train your dragon: Diverse augmentation towards generalizable
  dense retrieval.
\newblock In \emph{Findings of the Association for Computational Linguistics:
  {EMNLP}}, pages 6385--6400.

\bibitem[{Lin et~al.(2024)Lin, Chen, Chen, Shi, Lomeli, James, Rodriguez, Kahn,
  Szilvasy, Lewis et~al.}]{lin2024ra}
Xi~Victoria Lin, Xilun Chen, Mingda Chen, Weijia Shi, Maria Lomeli, Rich James,
  Pedro Rodriguez, Jacob Kahn, Gergely Szilvasy, Mike Lewis, et~al. 2024.
\newblock $\text{RA-DIT}$: Retrieval-augmented dual instruction tuning.
\newblock In \emph{International Conference on Learning Representations}.

\bibitem[{Liu et~al.(2024)Liu, Lin, Hewitt, Paranjape, Bevilacqua, Petroni, and
  Liang}]{liu2024lost}
Nelson~F Liu, Kevin Lin, John Hewitt, Ashwin Paranjape, Michele Bevilacqua,
  Fabio Petroni, and Percy Liang. 2024.
\newblock Lost in the middle: How language models use long contexts.
\newblock \emph{Transactions of the Association for Computational Linguistics},
  12:157--173.

\bibitem[{Luo et~al.(2023)Luo, Li, Haffari, and Pan}]{luo2023reasoning}
Linhao Luo, Yuan-Fang Li, Gholamreza Haffari, and Shirui Pan. 2023.
\newblock Reasoning on graphs: Faithful and interpretable large language model
  reasoning.
\newblock \emph{arXiv preprint arXiv:2310.01061}.

\bibitem[{Ma et~al.(2023)Ma, Wang, Yang, Wei, and Lin}]{ma2023fine}
Xueguang Ma, Liang Wang, Nan Yang, Furu Wei, and Jimmy Lin. 2023.
\newblock Fine-tuning llama for multi-stage text retrieval.
\newblock \emph{arXiv preprint arXiv:2310.08319}.

\bibitem[{Min et~al.(2019)Min, Chen, Zettlemoyer, and
  Hajishirzi}]{min2019knowledge}
Sewon Min, Danqi Chen, Luke Zettlemoyer, and Hannaneh Hajishirzi. 2019.
\newblock Knowledge guided text retrieval and reading for open domain question
  answering.
\newblock \emph{arXiv preprint arXiv:1911.03868}.

\bibitem[{Nguyen et~al.(2024)Nguyen, Luo, Shiri, Phung, Li, Vu, and
  Haffari}]{nguyen2024direct}
Minh-Vuong Nguyen, Linhao Luo, Fatemeh Shiri, Dinh Phung, Yuan-Fang Li,
  Thuy-Trang Vu, and Gholamreza Haffari. 2024.
\newblock Direct evaluation of chain-of-thought in multi-hop reasoning with
  knowledge graphs.
\newblock \emph{arXiv preprint arXiv:2402.11199}.

\bibitem[{Nogueira et~al.(2020)Nogueira, Jiang, Pradeep, and
  Lin}]{nogueira2020document}
Rodrigo~Frassetto Nogueira, Zhiying Jiang, Ronak Pradeep, and Jimmy Lin. 2020.
\newblock Document ranking with a pretrained sequence-to-sequence model.
\newblock In \emph{Findings of the Association for Computational Linguistics:
  {EMNLP}}, pages 708--718.

\bibitem[{Oguz et~al.(2022)Oguz, Chen, Karpukhin, Peshterliev, Okhonko,
  Schlichtkrull, Gupta, Mehdad, and Yih}]{oguz2022unik}
Barlas Oguz, Xilun Chen, Vladimir Karpukhin, Stan Peshterliev, Dmytro Okhonko,
  Michael~Sejr Schlichtkrull, Sonal Gupta, Yashar Mehdad, and Scott Yih. 2022.
\newblock $\text{UniK-QA}$: Unified representations of structured and
  unstructured knowledge for open-domain question answering.
\newblock In \emph{Findings of the Association for Computational Linguistics},
  pages 1535--1546.

\bibitem[{Petroni et~al.(2020)Petroni, andAleksandra Piktus, Rockt{\"{a}}schel,
  Wu, Miller, and Riedel}]{petroni2020how}
Fabio Petroni, Patrick S. H.~Lewis andAleksandra Piktus, Tim Rockt{\"{a}}schel,
  Yuxiang Wu, Alexander~H. Miller, and Sebastian Riedel. 2020.
\newblock How context affects language models' factual predictions.
\newblock In \emph{Conference on Automated Knowledge Base Construction}.

\bibitem[{Ram et~al.(2023)Ram, Levine, Dalmedigos, Muhlgay, Shashua,
  Leyton-Brown, and Shoham}]{ram2023context}
Ori Ram, Yoav Levine, Itay Dalmedigos, Dor Muhlgay, Amnon Shashua, Kevin
  Leyton-Brown, and Yoav Shoham. 2023.
\newblock In-context retrieval-augmented language models.
\newblock \emph{Transactions of the Association for Computational Linguistics},
  11:1316--1331.

\bibitem[{Ramesh et~al.(2023)Ramesh, Sreedhar, and Hu}]{ramesh2023single}
Gowtham Ramesh, Makesh~Narsimhan Sreedhar, and Junjie Hu. 2023.
\newblock Single sequence prediction over reasoning graphs for multi-hop
  $\text{QA}$.
\newblock In \emph{Proceedings of the 61st Annual Meeting of the Association
  for Computational Linguistics (Volume 1: Long Papers)}, pages 11466--11481.

\bibitem[{Rubin et~al.(2022)Rubin, Herzig, and Berant}]{rubin2022learning}
Ohad Rubin, Jonathan Herzig, and Jonathan Berant. 2022.
\newblock Learning to retrieve prompts for in-context learning.
\newblock In \emph{Proceedings of the 2022 Conference of the North American
  Chapter of the Association for Computational Linguistics}, pages 2655--2671.

\bibitem[{Sanmartin(2024)}]{sanmartin2024kg}
Diego Sanmartin. 2024.
\newblock $\text{KG-RAG}$: Bridging the gap between knowledge and creativity.
\newblock \emph{arXiv preprint arXiv:2405.12035}.

\bibitem[{Shi et~al.(2023{\natexlab{a}})Shi, Chen, Misra, Scales, Dohan, Chi,
  Sch{\"a}rli, and Zhou}]{shi2023large}
Freda Shi, Xinyun Chen, Kanishka Misra, Nathan Scales, David Dohan, Ed~H Chi,
  Nathanael Sch{\"a}rli, and Denny Zhou. 2023{\natexlab{a}}.
\newblock Large language models can be easily distracted by irrelevant context.
\newblock In \emph{International Conference on Machine Learning}, pages
  31210--31227. PMLR.

\bibitem[{Shi et~al.(2023{\natexlab{b}})Shi, Min, Yasunaga, Seo, James, Lewis,
  Zettlemoyer, and Yih}]{shi2023replug}
Weijia Shi, Sewon Min, Michihiro Yasunaga, Minjoon Seo, Rich James, Mike Lewis,
  Luke Zettlemoyer, and Wen-tau Yih. 2023{\natexlab{b}}.
\newblock $\text{REPLUG}$: Retrieval-augmented black-box language models.
\newblock \emph{arXiv preprint arXiv:2301.12652}.

\bibitem[{Trivedi et~al.(2022)Trivedi, Balasubramanian, Khot, and
  Sabharwal}]{trivedi2022musique}
Harsh Trivedi, Niranjan Balasubramanian, Tushar Khot, and Ashish Sabharwal.
  2022.
\newblock $\text{MuSiQue}$: Multihop questions via single-hop question
  composition.
\newblock \emph{Trans. Assoc. Comput. Linguistics}, 10:539--554.

\bibitem[{Trivedi et~al.(2023)Trivedi, Balasubramanian, Khot, and
  Sabharwal}]{trivedi2023interleaving}
Harsh Trivedi, Niranjan Balasubramanian, Tushar Khot, and Ashish Sabharwal.
  2023.
\newblock Interleaving retrieval with chain-of-thought reasoning for
  knowledge-intensive multi-step questions.
\newblock In \emph{Proceedings of the 61st Annual Meeting of the Association
  for Computational Linguistics (Volume 1: Long Papers)}, pages 10014--10037.

\bibitem[{Vrande{\v{c}}i{\'c} and Kr{\"o}tzsch(2014)}]{vrandevcic2014wikidata}
Denny Vrande{\v{c}}i{\'c} and Markus Kr{\"o}tzsch. 2014.
\newblock Wikidata: a free collaborative knowledgebase.
\newblock \emph{Communications of the ACM}, 57(10):78--85.

\bibitem[{Wang et~al.(2022)Wang, Yang, Huang, Jiao, Yang, Jiang, Majumder, and
  Wei}]{wang2022text}
Liang Wang, Nan Yang, Xiaolong Huang, Binxing Jiao, Linjun Yang, Daxin Jiang,
  Rangan Majumder, and Furu Wei. 2022.
\newblock Text embeddings by weakly-supervised contrastive pre-training.
\newblock \emph{arXiv preprint arXiv:2212.03533}.

\bibitem[{Wang et~al.(2023)Wang, Yang, Huang, Yang, Majumder, and
  Wei}]{wang2023improving}
Liang Wang, Nan Yang, Xiaolong Huang, Linjun Yang, Rangan Majumder, and Furu
  Wei. 2023.
\newblock Improving text embeddings with large language models.
\newblock \emph{arXiv preprint arXiv:2401.00368}.

\bibitem[{Wei et~al.(2022)Wei, Tay, Bommasani, Raffel, Zoph, Borgeaud,
  Yogatama, Bosma, Zhou, Metzler, Chi, Hashimoto, Vinyals, Liang, Dean, and
  Fedus}]{wei2022jason}
Jason Wei, Yi~Tay, Rishi Bommasani, Colin Raffel, Barret Zoph, Sebastian
  Borgeaud, Dani Yogatama, Maarten Bosma, Denny Zhou, Donald Metzler, Ed~H.
  Chi, Tatsunori Hashimoto, Oriol Vinyals, Percy Liang, Jeff Dean, and William
  Fedus. 2022.
\newblock Emergent abilities of large language models.
\newblock \emph{Transactions on Machine Learning Research}, 2022.

\bibitem[{Wei et~al.(2023)Wei, Cui, Cheng, Wang, Zhang, Huang, Xie, Xu, Chen,
  Zhang et~al.}]{wei2023zero}
Xiang Wei, Xingyu Cui, Ning Cheng, Xiaobin Wang, Xin Zhang, Shen Huang, Pengjun
  Xie, Jinan Xu, Yufeng Chen, Meishan Zhang, et~al. 2023.
\newblock Zero-shot information extraction via chatting with chatgpt.
\newblock \emph{arXiv preprint arXiv:2302.10205}.

\bibitem[{Xiao et~al.(2023)Xiao, Liu, Zhang, and Muennighof}]{xiao2023c}
Shitao Xiao, Zheng Liu, Peitian Zhang, and Niklas Muennighof. 2023.
\newblock C-pack: Packaged resources to advance general chinese embedding.
\newblock \emph{arXiv preprint arXiv:2309.07597}.

\bibitem[{Yang et~al.(2018)Yang, Qi, Zhang, Bengio, Cohen, Salakhutdinov, and
  Manning}]{yang2018hotpotqa}
Zhilin Yang, Peng Qi, Saizheng Zhang, Yoshua Bengio, William~W. Cohen, Ruslan
  Salakhutdinov, and Christopher~D. Manning. 2018.
\newblock $\text{HotpotQA}$: {A} dataset for diverse, explainable multi-hop
  question answering.
\newblock In \emph{Proceedings of the 2018 Conference on Empirical Methods in
  Natural Language Processing}, pages 2369--2380.

\bibitem[{Yavuz et~al.(2022)Yavuz, Hashimoto, Zhou, Keskar, and
  Xiong}]{yavuz2022modeling}
Semih Yavuz, Kazuma Hashimoto, Yingbo Zhou, Nitish~Shirish Keskar, and Caiming
  Xiong. 2022.
\newblock Modeling multi-hop question answering as single sequence prediction.
\newblock In \emph{Proceedings of the 60th Annual Meeting of the Association
  for Computational Linguistics}, pages 974--990.

\bibitem[{Yoran et~al.(2024)Yoran, Wolfson, Ram, and Berant}]{yoran2024making}
Ori Yoran, Tomer Wolfson, Ori Ram, and Jonathan Berant. 2024.
\newblock Making retrieval-augmented language models robust to irrelevant
  context.
\newblock In \emph{International Conference on Learning Representations}.

\bibitem[{Yu et~al.(2022)Yu, Zhu, Fang, Yu, Wang, Xu, Ren, Yang, and
  Zeng}]{yu2022kg}
Donghan Yu, Chenguang Zhu, Yuwei Fang, Wenhao Yu, Shuohang Wang, Yichong Xu,
  Xiang Ren, Yiming Yang, and Michael Zeng. 2022.
\newblock $\text{KG-FiD}$: Infusing knowledge graph in fusion-in-decoder for
  open-domain question answering.
\newblock In \emph{Proceedings of the 60th Annual Meeting of the Association
  for Computational Linguistics}, pages 4961--4974.

\bibitem[{Zhang and Soh(2024)}]{zhang2024extract}
Bowen Zhang and Harold Soh. 2024.
\newblock Extract, define, canonicalize: An llm-based framework for knowledge
  graph construction.
\newblock \emph{arXiv preprint arXiv:2404.03868}.

\bibitem[{Zhang et~al.(2022)Zhang, Zhang, Yu, Tang, Tang, Li, and
  Chen}]{zhang2022subgraph}
Jing Zhang, Xiaokang Zhang, Jifan Yu, Jian Tang, Jie Tang, Cuiping Li, and Hong
  Chen. 2022.
\newblock Subgraph retrieval enhanced model for multi-hop knowledge base
  question answering.
\newblock In \emph{Proceedings of the 60th Annual Meeting of the Association
  for Computational Linguistics (Volume 1: Long Papers)}, pages 5773--5784.

\bibitem[{Zhou et~al.(2020)Zhou, Shi, Huang, and Zhu}]{zhou2020knowledge}
Mantong Zhou, Zhouxing Shi, Minlie Huang, and Xiaoyan Zhu. 2020.
\newblock Knowledge-aided open-domain question answering.
\newblock \emph{arXiv preprint arXiv:2006.05244}.

\end{thebibliography}

\begin{algorithm*}[tb]
\DontPrintSemicolon
\LinesNumbered
\SetKwInOut{Input}{Input}\SetKwInOut{Output}{Output}
\Input{question $q$, a set of retrieved documents $\mathcal{D}_q = \{d_1, d_2, \dots, d_N\}$, maximum reasoning chain length $L$, number of candidate triples $K$, number of reasoning chains $R$}
    \tcc{Part I: Knowledge Graph Generation.}
    $\mathcal{G}_q = \{ \}$; \# Initialise KG \\ 
    \For{$d_i \in \mathcal{D}_q$}{
        $\mathcal{G}_{q, d_i}=\textit{KG\_Generator}(d_i)$; \\ 
        $\mathcal{G}_q = \mathcal{G}_q.\text{add}(\mathcal{G}_{q, d_i})$; \\
    }
    
    \tcc{Part II: Reasoning Chain Construction.}
    $z = $[[] for \_ in range($R$)]; $c = $[[q] for \_ in range($R$)]; $s = $[[1.0] for \_ in range($R$)]; \\ 
    \For{$i=1, 2, \dots, L$}{
        $c\_z, c\_c, c\_s = $[], [], []; \\ 
        \For{$r = 1, 2, \dots, R$}{
            $\hat{\mathcal{G}}_i = Triple\_Scorer(c[r], \mathcal{G}_q, K)$; \\
            $p(z_i | q, z_{<i}, \hat{\mathcal{G}}_i) = Triple\_Selector(c[r], \hat{\mathcal{G}}_i)$; \\
            \For{$z_{i,j} \in$ \textit{argmax$_b$}$(p(z_i | q, z_{<i}, \hat{\mathcal{G}}_i))$}{
                $c\_z$.append($z$[$r$] + [$z_{i,j}$]); \\
                $c\_c$.append($c$[$r$] + [$z_{i,j}$]); \\
                $c\_s$.append($s$[$r$]$*p(z_i = z_{i,j} | q, z_{<i}, \hat{\mathcal{G}}_i)$); \\ 
            }
        }
        indices = \textit{argmax$_R$}($c\_s$); $z = c\_z$[indices]; $c = c\_c$[indices]; \\
    }

\Output{Reasoning Chains $z$.}
    \caption{Knowledge Triple-Grounded Reasoning Chains Construction of \OURS{}.}
    \label{alg:trace}
\end{algorithm*}

\begin{table}[tb]
\centering
\resizebox{0.48\textwidth}{!}{
\begin{tabular}{cc}
\toprule
\textbf{Model} & \textbf{Huggingface Checkpoint} \\
\midrule
Contriever & {\tt contriever} \\ 
DRAGON+ & {\tt dragon-plus} \\ 
E5 & {\tt e5-large-v2} \\ 
E5-Mistral & {\tt e5-mistral-7b-instruct} \\ 
monoT5 & {\tt monot5-large-msmarco} \\ 
BGE & {\tt bge-reranker-large} \\ 
RankLLaMA & {\tt rankllama-v1-7b-lora-passage} \\ 
\bottomrule
\end{tabular}
} 
\vspace{-0.5em}
\caption{The specific retrieval models used in our experiments.}
\label{table:retrieval_models}
\vspace{-1.0em}
\end{table}

\appendix

\section{\jy{Reasoning Chain Construction Algorithm and Complexity Analysis}}
\label{app:algorithm}

The algorithm for the reasoning chain construction process is detailed in Algorithm~\ref{alg:trace}, which illustrates the construction of $R$ reasoning chains from a set of documents $\mathcal{D}_q$ to answer a question $q$. For clarity, we omit the ``adaptive chain termination'' strategy in the algorithm. However, it can be easily incorporated into the algorithm if needed. 

The computational complexity of constructing the reasoning chains is $\mathcal{O}(N+LR)$, where $N$ is the number of documents in $\mathcal{D}_q$ and $L$ denotes the length of the reasoning chains. This complexity applies when the adaptive chain termination strategy is not used. Incorporating this strategy can improve the actual efficiency further. 

\begin{figure*}[tb]
\centering
\begin{small}

    \begin{tcolorbox}[colframe=black!75!white, coltitle=black, colback=white, boxrule=0.3mm, boxsep=0.5mm, width=\textwidth]
    \textcolor{gray}{Instruction:}\vspace{0.2em}

    Given a title and a text, extract all the knowledge triples in the form of $\langle$title; relation; tail entity$\rangle$, where title is the provided title, tail entity is a phrase in the text and relation denotes a description of the relation between the title and the tail entity. 
    \vspace{\baselineskip}
    
    \textcolor{gray}{Demonstrations:}\vspace{0.2em} \\
    \textbf{Title}: Albert Einstein\\
    \textbf{Text}: Albert Einstein (14 March 1879-18 April 1955) was a German-born theoretical physicist. \\
    \textbf{Knowledge Triples}: \\
    $\langle$Albert Einstein; date of birth; 14 March 1879$\rangle$\\
    $\langle$Albert Einstein; date of death; 18 April 1955$\rangle$\\
    $\langle$Albert Einstein; place of birth; German$\rangle$\\
    $\langle$Albert Einstein; occupation; theoretical physicist$\rangle$ \\
    $\dots$ \vspace{\baselineskip}
    
    \textcolor{gray}{Input Document:}\vspace{0.2em} \\
    \textbf{Title}: Kelie McIver \\
    \textbf{Text}: Kelie McIver is a Kansas-born actress and singer who has played classical stage roles such as Lady Macbeth and Nurse in ``Romeo \& Juliet'' for Kingsmen Shakespeare Festival. \\
    \textbf{Knowledge Triples}: 
    
    \end{tcolorbox}
    \vspace{-1.0em}
    \caption{Prompt for generating knowledge triples from a document.}
    \label{fig:kg_generation_prompt}
    \vspace{-0.5em}
\end{small}
\end{figure*}

\begin{figure*}[!ht]
\centering
\begin{small}
    \begin{tcolorbox}[colframe=black!75!white, coltitle=black, colback=white, boxrule=0.3mm, boxsep=0.5mm, width=\textwidth]
    \textcolor{gray}{Instruction:}\vspace{0.2em}
    
    Select the next knowledge triple that extends an existing set of knowledge triples to form a coherent reasoning path capable of answering a specified question. If the current reasoning path is sufficient to answer the question, simply output A. Please only output the choice for the next knowledge triple. \vspace{\baselineskip}
    
    The following are some examples of coherent reasoning paths capable of answering the specified question and how the $l$-th knowledge triples in these paths are selected: \vspace{\baselineskip}
    
    \textcolor{gray}{Demonstrations:}\vspace{0.2em} 
    
    \textbf{coherent reasoning path}: $\langle$A Girl's Gotta Do (What a Girl's Gotta Do); artist; American country music artist Mindy McCready$\rangle$, $\langle$Mindy McCready; fifth album; I'm Still Here$\rangle$ 
    
    \textbf{question}: What is the 5th studio album released by the singer of "A Girl's Gotta Do (What a Girl's Gotta Do)"? \vspace{\baselineskip}
    
    \textbf{The $l$-th triple in the reasoning path is selected as:} \\
    \textbf{existing knowledge triples}: \textcolor{gray}{[Previously selected triples in the form of $\langle\cdot\rangle$, $\langle\cdot \rangle$, \dots]} \\
    \textbf{question}: What is the 5th studio album released by the singer of "A Girl's Gotta Do (What a Girl's Gotta Do)"? \\
    \textbf{candidate knowledge triples}:\\
    A. no need for additional knowledge triples \\
    B. $\langle$A Girl's Gotta Do (What a Girl's Gotta Do); artist; American country music artist Mindy McCready$\rangle$ \\
    C. $\langle$A Girl's Gotta Do (What a Girl's Gotta Do); release date; February 1997$\rangle$ \\
    D. $\langle$Ten Thousand Angels; fourth single; "A Girl's Gotta Do (What a Girl's Gotta Do)"$\rangle$ \\
    E. $\langle$A Girl's Gotta Do (What a Girl's Gotta Do); songwriters; Robert Byrne and Rick Bowles$\rangle$ \\
    \textbf{the next possible triple is}:B \\
    $\dots$ 
    \vspace{\baselineskip}
    
    \textcolor{gray}{Input Document:}\vspace{0.2em} \\
    \textbf{The $l$-th triple in the reasoning path is selected as}:\\
    \textbf{existing knowledge triples}: \textcolor{gray}{[Previously selected triples in the form of $\langle\cdot\rangle$, $\langle\cdot \rangle$, \dots]} \\
    \textbf{question}: Are Ellen Glasgow and Günter Grass both novelists? \\
    \textbf{candidate knowledge triples}: \\
    A. no need for additional knowledge triples \\
    B. $\langle$Ellen Glasgow; occupation; novelist$\rangle$\\
    C. $\langle$Virginia (novel); author; Ellen Glasgow$\rangle$\\
    D. $\langle$Ellen Glasgow; full name; Ellen Anderson Gholson Glasgow$\rangle$\\
    E. $\langle$Günter Grass; occupation; novelist, poet, playwright, illustrator, graphic artist, sculptor$\rangle$\\
    \textbf{the next possible triple is}:
    
    \end{tcolorbox}
    \vspace{-1.0em}
    \caption{Prompt for selecting the $i$-th triples of the reasoning chains.}
    \label{fig:triple_selection_prompt}
    \vspace{-1.0em}
\end{small}
\end{figure*}

\begin{figure*}[tb]
    \centering
\begin{small}
    \begin{tcolorbox}[colframe=black!75!white, coltitle=black, colback=white, boxrule=0.3mm, boxsep=0.5mm, width=\textwidth]
    \textcolor{gray}{Instruction:}\vspace{0.2em}

    Given some contexts and a question, please only output the answer to the question. \\
    \textbf{context}: \\
    ... ({reasoning chains / documents}) \\
    \textbf{the correct answer is}: 
    
    \end{tcolorbox}
    \caption{Prompt for generating answers to the questions based on given context.}
    \label{fig:answer_generation_prompt}
\end{small}
\end{figure*}

\begin{table*}[tb]
\centering
\resizebox{0.7\textwidth}{!}{
\begin{tabular}{lcccccc}
\toprule
& \multicolumn{2}{c}{\textbf{HotPotQA}} & \multicolumn{2}{c}{\textbf{2WikiMultiHopQA}} & \multicolumn{2}{c}{\textbf{MuSiQue}} \\
\cmidrule(lr){2-3} \cmidrule(lr){4-5} \cmidrule(lr){6-7}
& Dev. & Test & \textcolor{white}{00}Dev. & Test & Dev. & Test \\
\midrule
\noalign{\vskip 0.5ex} \multicolumn{7}{l}{\textbf{\textit{Statistics of Multi-Hop QA datasets}}} \vspace{0.3em} \\
\textbf{\# Questions} & 500 &7,405 & \textcolor{white}{00}500 &12,576 & 500 &2,417 \\
\textbf{\# Documents per Question} & 10 & 10 & \textcolor{white}{00}10 & 10 & 20 & 20 \\ 
\textbf{Avg. Document Error Rate (\%)} & 79.52 & 79.66 & \textcolor{white}{00}75.68 & 75.62 & 88.19 & 86.74 \\ 

\cmidrule(lr){1-7}\noalign{\vskip 0.5ex}
\multicolumn{7}{l}{\textit{\textbf{Statistics of the Generated KG}}}\vspace{0.3em} \\ 
\textbf{Avg. \# Entities per Question} & 83.21 & 83.63 & \textcolor{white}{00}65.67 & 71.21 & 159.87 & 164.03 \\ 
\textbf{Avg. \# Triples per Question} & 79.06 & 79.04 & \textcolor{white}{00}62.75 & 67.46 & 150.53 & 154.70  \\ 
\textbf{Avg. Density per Question (\%)} & 1.23 & 1.20 & \textcolor{white}{00}1.57 & 1.42 & 0.61 & 0.59 \\ 

\cmidrule(lr){1-7}\noalign{\vskip 0.5ex}
\multicolumn{7}{l}{\textit{\textbf{Statistics of the Constructed Reasoning Chains}}}\vspace{0.3em} \\
\textbf{Avg. Reasoning Chain Length} & 3.15 & 3.16 & \textcolor{white}{00}3.14 & 3.36 & 3.14 & 3.17  \\
\textbf{Avg. \# Relevant Documents per Question} & 2.63 & 2.57 & \textcolor{white}{00}2.73 & 2.82 & 2.70 & 2.84  \\ 
\textbf{Avg. Document Error Rate (\%)} & 22.78 & 21.06 & \textcolor{white}{00}12.64 & 14.82 & 28.48 & 29.76 \\ 
\bottomrule    
\end{tabular}
}
\caption{Statistics of the generated KGs and reasoning chains, where ``Avg. \# Relevant Documents per Question'' denotes the average number of relevant documents identified with reasoning chains and ``Avg. Document Error rate (\%)'' denotes the average percentage of documents that are irrelevant to the questions.}
\label{table:statistics}
\end{table*}

\begin{table}[tb]
\centering
\resizebox{0.48\textwidth}{!}{
\begin{tabular}{lcccccc}
\toprule
\multirow{2}{*}{\textbf{Model}} & \multicolumn{2}{c}{\textbf{HotPotQA}} & \multicolumn{2}{c}{\textbf{2WikiMultiHopQA}} & \multicolumn{2}{c}{\textbf{MuSiQue}}  \\
\cmidrule(lr){2-3} \cmidrule(lr){4-5} \cmidrule(lr){6-7} 
& \textbf{EM} & \textbf{F1} & \textcolor{white}{00}\textbf{EM} & \textbf{F1} & \textbf{EM} & \textbf{F1}  \\
\midrule

\textbf{\OURS{}-Doc} & \textbf{69.20}\textcolor{white}{$^*$} & \textbf{78.46}\textcolor{white}{$^*$} & \textcolor{white}{00}\textbf{68.00}\textcolor{white}{$^*$} & \textbf{75.96}\textcolor{white}{$^*$} & \textbf{41.00}\textcolor{white}{$^*$} & \textbf{47.09}\textcolor{white}{$^*$} \\ 

\cmidrule(lr){1-7}\noalign{\vskip 0.5ex}
\multicolumn{7}{l}{\textit{\textbf{Effectiveness of KG Generator}}}\vspace{0.3em} \\ 
\textbf{w. sentences} & 67.20{$^*$} & 77.56{$^*$} & \textcolor{white}{00}65.00{$^*$} & 73.06{$^*$} & 33.60{$^*$} & 39.62{$^*$} \\
\textbf{w. documents} & 66.20{$^*$} & 76.73{$^*$} & \textcolor{white}{00}57.60{$^*$} & 67.83{$^*$} & 30.80{$^*$} & 36.13{$^*$} \\

\cmidrule(lr){1-7}\noalign{\vskip 0.5ex}
\multicolumn{7}{l}{\textit{\textbf{Effectiveness of Reasoning Chain Constructor}}}\vspace{0.3em} \\ 
\textbf{w. Top-10 Triples} & 63.80{$^*$} & 73.30{$^*$} & \textcolor{white}{00}49.80{$^*$} & 56.27{$^*$} & 32.40{$^*$} & 38.11{$^*$} \\
\textbf{w. Top-20 Triples} & 62.60{$^*$} & 73.60{$^*$} & \textcolor{white}{00}52.40{$^*$} & 59.97{$^*$} & 30.80{$^*$} & 37.48{$^*$} \\

\cmidrule(lr){1-7}\noalign{\vskip 0.5ex}
\multicolumn{7}{l}{\textit{\textbf{Effectiveness of Triple Ranker}}}\vspace{0.3em} \\ 
\textbf{w/o Triple Ranker} & 63.40{$^*$} & 73.79{$^*$} & \textcolor{white}{00}62.20{$^*$} & 71.59{$^*$} & 29.40{$^*$} & 35.18{$^*$} \\ 
\textbf{w. DRAGON+} & 64.60{$^*$} & 74.92{$^*$} & \textcolor{white}{00}65.00{$^*$} & 73.67{$^*$} & 36.20{$^*$} & 41.76{$^*$} \\
\textbf{w. E5} & 67.80{$^*$} & 77.82{$^*$} & \textcolor{white}{00}67.20\textcolor{white}{$^*$} & 75.27\textcolor{white}{$^*$} & 39.00\textcolor{white}{$^*$} & 45.96\textcolor{white}{$^*$} \\

\cmidrule(lr){1-7}\noalign{\vskip 0.5ex}
\multicolumn{7}{l}{\textit{\textbf{Effectiveness of Triple Selector}}}\vspace{0.3em} \\ 
\textbf{w/o Triple Selector} &59.80{$^*$} & 69.38{$^*$} & \textcolor{white}{00}49.20{$^*$} & 55.39{$^*$} & 30.00{$^*$} & 35.56{$^*$} \\ 
\textbf{w. Mistral} &64.40{$^*$} & 74.91{$^*$} & \textcolor{white}{00}65.60{$^*$} & 74.13\textcolor{white}{$^*$} & 35.60{$^*$} & 42.65{$^*$} \\ 
\textbf{w. Gemma} &63.60{$^*$} & 72.49{$^*$} & \textcolor{white}{00}63.80{$^*$} & 72.13{$^*$} & 31.00{$^*$} & 37.17{$^*$} \\ 

\bottomrule
\end{tabular}
} 
\caption{Performance (\%) of \OURS{}-Doc and its variants on the development sets of three QA datasets, where $^*$ denotes p<0.05 compared with \OURS{}-Doc.}
\label{table:ablation_study_use_document}
\end{table}

\section{Prompts}
In this section, we present the prompts used in our \OURS{}. Specifically, the prompt for generating KG triples is detailed in \S~\ref{app:prompts_for_generating_kg_triples}, the prompt for the triple selector is outlined in \S~\ref{app:prompts_for_generating_reasoning_chains}, and the prompt for generating answers in the reader is introduced in \S~\ref{app:prompts_for_generating_answers}. 
Finally, we provide some examples of demonstrations in \S~\ref{app:examples_of_labeled_data}.

\subsection{Prompt for Generating KG Triples}
\label{app:prompts_for_generating_kg_triples}
The KG generator independently generates KG triples for each document within $\mathcal{D}_q$. The prompt used for the KG generator to generate KG triples from a document is provided in Figure~\ref{fig:kg_generation_prompt}. 
Specifically, the prompt comprises three parts: instruction, demonstrations and input document. The instruction defines the task of generating KG triples from a document. The demonstrations are examples of documents and their corresponding KG triples. The input document is the document from which we expect the KG generator to generate KG triples. 
The output of the KG generator is all the possible KG triples identified within the input document.

\subsection{Prompt for Triple Selector}
\label{app:prompts_for_generating_reasoning_chains}

The triple selector aims to select a triple one by one to construct reasoning chains. 
The prompt used by the triple selector to select the $i$-th triple is provided in Figure~\ref{fig:triple_selection_prompt}. Specifically, the prompt consists of three parts: instruction, demonstrations and inputs. 
The instruction defines the task of selecting a triple to form coherent reasoning chains. 
The demonstrations are examples of complete reasoning chains and how the $i$-th triples in these chains are selected. The inputs consist of the question $q$, existing knowledge triples $z_{<i}$ and the candidate set $\hat{\mathcal{G}}_i$. 
Moreover, the triple selector is instructed to output only the option of the selected triples, such as ``A'', ``B'', ect.

\subsection{Prompt for Generating Answers}
\label{app:prompts_for_generating_answers}

In our experiments, we use in-context learning to prompt the reader to generate answers in a zero-shot setting. The prompt used for the answer generation is provided in Figure~\ref{fig:answer_generation_prompt}. Moreover, the reader is instructed to output only the answer to the question based on the given context, such as reasoning chains or documents. 

\subsection{Examples of Labeled Data for KG Generator and Triple Selector}
\label{app:examples_of_labeled_data}
The examples of labelled data for the KG generator and the triple selector are provided in Table~\ref{table:examples_for_kg_generator} and Table~\ref{table:examples_for_triple_selector}, respectively.

\section{Experimental Details}
\jy{Due to the space limit, we provide additional experimental details in this section, which are complementary to \S~\ref{subsec:experimental_setup} in the main text. Specifically, we introduce further details about the documents and the data splits of the experimental datasets in \S~\ref{app:datasets}. We then detail the specific parameterisations of the baselines in \S~\ref{app:baselines}. Moreover, we provide the statistics and analyses of the generated KGs and reasoning chains in \S~\ref{app:statistics_kg_chains}. Finally, we include additional implementation details, such as prompt demonstrations and hyperparameters, in \S~\ref{app:implementation}.}

\subsection{Datasets}
\label{app:datasets}
We use three multi-hop QA datasets: HotPotQA, 2WikiMultiHopQA, and MuSiQue. Each question in these datasets is provided with a set of documents retrieved from Wikipedia, which include both relevant and irrelevant documents. These documents are randomly shuffled in the original datasets. Additionally, the datasets provide annotations indicating which documents are relevant to each question. Since \OURS{} aims to identify supporting evidence within the retrieved documents, we directly use these documents as inputs and perform multi-hop reasoning over these documents to generate answers. 
Since the test sets of these datasets are not publicly available, we follow~\citet{ramesh2023single} and report the performance on the original development sets, which contain 7,405 questions for HotPotQA, 12,576 questions for 2WikiMultiHopQA, and 2,417 questions for MuSiQue. We randomly sample 500 questions from the training set of each dataset to create our development sets for hyperparameter tuning. 

\subsection{Baselines}
\label{app:baselines}
In order to evaluate the effectiveness of \OURS{} in identifying supporting evidence, we mainly compare with baselines that use different methods to identify relevant documents from the document set $\mathcal{G}_q$. Once the relevant documents are obtained, they are used as input to the same reader as \OURS{} to generate answers. 
Specifically, we compare with baselines from the following categories:

\vspace{0.5em} \noindent \textbf{w/o documents}: In this baseline, we remove all the documents and use only the questions as inputs for the reader to generate answers. 

\vspace{0.5em} \noindent \textbf{w. all documents}: In this baseline, we use both the questions and all the documents as inputs for the reader (i.e., the vanilla RAG). Note that the documents provided by the datasets, i.e., $\mathcal{G}_q$, are already randomly shuffled. We do not perform any ranking on these documents; instead, we directly concatenate all the documents in their original order. 

\vspace{0.5em} \noindent \textbf{bi-encoders/cross-encoders}: To identify relevant documents within $\mathcal{G}_q$ with retrieval models, such as bi-encoders and cross-encoders, we use these models to estimate the relevance scores between a question and all the documents within $\mathcal{G}_q$. We then rank these documents in descending order based on the estimated relevance scores. The top-$M$ documents are considered relevant and used as inputs for the reader, where $M$ is selected from $\{1, 2, \dots, N\}$ and is set to the number of documents that results in the best performance of the reader on the development set of each dataset. The checkpoints we use for different retrieval models are in Table~\ref{table:retrieval_models}.

\vspace{0.5em} \noindent \textbf{IRCoT}: 
IRCoT was originally proposed to leverage chain-of-thought (CoT) sentences for both document retrieval and answer generation. Here, we use it solely to retrieve relevant documents from $\mathcal{G}_q$. For a fair comparison with \OURS{}, we use the same {\tt LLaMA3-8B-Instruct} model to generate CoT sentences. 
Following the original methodology, we alternate between CoT sentence generation and document retrieval to retrieve a set of relevant documents. 
The resulting documents are used as input to the reader to generate answers. 

\subsection{Statistics and Analyses of the Generated KGs and Reasoning Chains}
\label{app:statistics_kg_chains}
The statistics of the generated KGs and reasoning chains for our experimental datasets are provided in Table~\ref{table:statistics}. 
Specifically, we report the average number of entities and triples, as well as the average density in the KGs. For example, in the test set of the HotPotQA dataset, the average number of entities and triples is $83.21$ and $79.06$, respectively. The corresponding average density is 1.20\%, indicating that the KGs are highly sparse. The same results can also be observed in other datasets. This is because it is challenging to infer relationships between entities from different documents, or there may be no meaningful relationships between these entities, resulting in fewer connections and a lower overall density in the KGs. 

Moreover, the results in Table~\ref{table:statistics} show that the average length of reasoning chains is relatively small, approximately 3 across all datasets. Leveraging these reasoning chains to identify a subset of relevant documents from $\mathcal{D}_q$ results in an average of around 2.6 documents across all datasets. Despite the small number of documents, the error rates of these documents are significantly lower than those of the original documents $\mathcal{D}_q$, e.g., $21.06\%$ v.s. $79.66\%$ on the test set of HotPotQA. These results demonstrate the effectiveness of leveraging reasoning chains to identify a subset of relevant documents from $\mathcal{D}_q$ while avoiding the introduction of irrelevant ones. This can also explain the superior performance of \OURS{}-Doc over the vanilla RAG model, as the documents identified by \OURS{}-Doc contain significantly less noise. 

\jy{Furthermore, due to the lack of annotated data, we are unable to quantitatively evaluate the effectiveness of the generated KGs and reasoning chains. Instead, we provide some qualitative results to assess their performance. Particularly, we provide some examples of the generated KGs and reasoning chains on HotPotQA dataset in Table~\ref{table:examples_of_generated_kgs} and Table~\ref{table:case_study_full_results}, respectively. The results in Table~\ref{table:examples_of_generated_kgs} show that our KG generate can produce high-quality knowledge triples. Moreover, almost all the generated knowledge triples are grounded in the documents, with minimal instances of hallucinations. This reliable grounding serves as a solid foundation for the subsequent reasoning, ensuring the accuracy and effectiveness of the reasoning process. Additionally, detailed analyses of the generated reasoning chains can be found in \S~\ref{app:case_study}. }

\subsection{Additional Implementation Details}
\label{app:implementation}

We conduct experiments in a zero-shot setting. The context, such as reasoning chains or documents, is prepended to questions, which are then passed to the reader for answer generation. We use greedy decoding to generate answers in the reader.  

Moreover, to obtain demonstrations for the KG generator, we manually label 50 documents from the training set of each dataset. Examples of the labelled data are provided in Appendix~\ref{app:examples_of_labeled_data}. The complete labelled data for each dataset are available in our Github repository. 
Following previous works~\cite{rubin2022learning,li2023unified}, we use E5-Mistral to retrieve the top three most similar documents from the labelled set as demonstrations when generating the KG for a document. 
Similarly, for the reasoning chain constructor, we manually label 20 questions from the training set of each dataset. The demonstrations for the reasoning chain constructor are obtained in a similar manner to the KG generator, using E5-Mistral to retrieve the top three most similar questions from the labelled set as the demonstrations. We empirically verify the effectiveness of the demonstrations in Appendix~\ref{app:effects_of_demonstrations}. 

Furthermore, throughout the experiments, we set the maximum reasoning chain length $L$ as 4, the number of reasoning chains $R$ as 5 and the number of beams $b$ as 5. The number of candidate triples $K$ is chosen from $\{15, 20, 25, 30\}$ to achieve the best performance on the development set.

\section{Additional Experimental Results and Analysis}
\label{app:additional_results_analysis}
In this section, we first introduce the overall performance of \OURS{} using different readers in \S~\ref{app:overall_performance_different_readers}. We then introduce the ablation studies for \OURS{}-Doc in \S~\ref{app:ablation_of_trace_doc} and the effects of maximum chain length $L$ on the MuSiQue dataset in \S~\ref{app:effects_of_l_on_musique}. Subsequently, we investigate the effects of the demonstrations, the effects of the number of reasoning chains $R$ and the effects of the number of candidate triples in \S~\ref{app:effects_of_demonstrations}, \S~\ref{app:effects_of_R} and \S~\ref{app:effects_of_K}, respectively. In addition, we examine the effectiveness of the adaptive chain termination strategy in \S~\ref{app:adaptive_chain_termination}. Finally, we provide the results and analyses of the case study in \S~\ref{app:case_study}. 

\subsection{Overall performance of \OURS{} using Different Readers}
\label{app:overall_performance_different_readers}

\begin{table*}[tb]
\centering
\resizebox{0.9\textwidth}{!}{
\begin{tabular}{lcccccccccccc}
\toprule
\multirow{2}{*}{Model} & \multicolumn{3}{c}{\textbf{HotPotQA}} & \multicolumn{3}{c}{\textbf{2WikiMultiHopQA}} & \multicolumn{3}{c}{\textbf{MuSiQue}}  \\
\cmidrule(lr){2-4} \cmidrule(lr){5-7} \cmidrule(lr){8-10} 
& \textbf{\# Tok} & \textbf{EM} & \textbf{F1} & \textbf{\# Tok} & \textbf{EM} & \textbf{F1} & \textbf{\# Tok} & \textbf{EM} & \textbf{F1} \\

\midrule\noalign{\vskip 0.1ex}
\multicolumn{10}{l}{\textit{\textbf{Mistral Reader with Context from Naive Baselines}}}\vspace{0.1ex} \\

\textbf{w/o documents} 
& \textcolor{white}{00}46 & 21.20\textcolor{white}{$^*$} & 28.68\textcolor{white}{$^*$}  
& \textcolor{white}{00}42 & 23.61\textcolor{white}{$^*$} & 27.20\textcolor{white}{$^*$}
& \textcolor{white}{00}46 & \textcolor{white}{$^*$}4.22\textcolor{white}{$^*$} & \textcolor{white}{$^*$}8.85\textcolor{white}{$^*$}
\\
\textbf{w. all documents} 
&1,627 &39.61\textcolor{white}{$^*$} &52.46\textcolor{white}{$^*$} 
& 1,190 & 30.42\textcolor{white}{$^*$} & 37.93\textcolor{white}{$^*$}
& 2,913 & 13.82\textcolor{white}{$^*$} & 20.02\textcolor{white}{$^*$}
\\
\cmidrule(lr){1-10}\noalign{\vskip 0.1ex}
\multicolumn{10}{l}{\textit{\textbf{Mistral Reader with Context from Bi-Encoders}}}\vspace{0.1ex} \\

\textbf{Contriever} 
&1,627 &39.69\textcolor{white}{$^*$} &52.46\textcolor{white}{$^*$} 
& 1,004 & 31.72\textcolor{white}{$^*$} & 40.04\textcolor{white}{$^*$} 
& \textcolor{white}{0}539 & 17.17\textcolor{white}{$^*$} & 24.32\textcolor{white}{$^*$} 
\\
\textbf{DRAGON+} 
&1,627 &39.59\textcolor{white}{$^*$} &52.47\textcolor{white}{$^*$} 
& \textcolor{white}{0}726 & 32.98\textcolor{white}{$^*$} & 40.98\textcolor{white}{$^*$}
& \textcolor{white}{0}520 & 15.72\textcolor{white}{$^*$} & 23.21\textcolor{white}{$^*$}
\\
\textbf{E5} 
&\textcolor{white}{0}456 &44.09\textcolor{white}{$^*$} &56.91\textcolor{white}{$^*$} 
& \textcolor{white}{0}486 & 34.52\textcolor{white}{$^*$} & 42.08\textcolor{white}{$^*$}
& \textcolor{white}{0}493 & 17.25\textcolor{white}{$^*$} & 24.72\textcolor{white}{$^*$}
\\
\textbf{E5-Mistral} 
&\textcolor{white}{0}463 &46.95\textcolor{white}{$^*$} &60.29\textcolor{white}{$^*$} 
& \textcolor{white}{0}504 & 35.96\textcolor{white}{$^*$} & 43.89\textcolor{white}{$^*$}
& \textcolor{white}{0}508 & 21.39\textcolor{white}{$^*$} & 28.01\textcolor{white}{$^*$}
\\
\cmidrule(lr){1-10}\noalign{\vskip 0.1ex}
\multicolumn{10}{l}{\textit{\textbf{Mistral Reader with Context from Cross-Encoders}}}\vspace{0.1ex} \\
\textbf{monoT5} 
&\textcolor{white}{0}784 &42.16\textcolor{white}{$^*$} &55.48\textcolor{white}{$^*$} 
& \textcolor{white}{0}506 & 35.31\textcolor{white}{$^*$} & 43.31\textcolor{white}{$^*$}
& \textcolor{white}{0}506 & 19.90\textcolor{white}{$^*$} & 27.56\textcolor{white}{$^*$}
\\
\textbf{BGE} 
&\textcolor{white}{0}472 &47.40\textcolor{white}{$^*$} &61.09\textcolor{white}{$^*$} 
& \textcolor{white}{0}534 & 35.03\textcolor{white}{$^*$} & 43.49\textcolor{white}{$^*$}
& \textcolor{white}{0}525 & 22.22\textcolor{white}{$^*$} & 30.38\textcolor{white}{$^*$}
\\
\textbf{RankLLaMA} 
&\textcolor{white}{0}484 &42.97\textcolor{white}{$^*$} & 55.67\textcolor{white}{$^*$} 
& \textcolor{white}{0}525 & 37.11\textcolor{white}{$^*$} & 45.37\textcolor{white}{$^*$}
& \textcolor{white}{0}539 & 18.45\textcolor{white}{$^*$} & 25.92\textcolor{white}{$^*$}
\\
\cmidrule(lr){1-10}
\noalign{\vskip 0.1ex}
\multicolumn{10}{l}{\textit{\textbf{Mistral Reader with Context from Chain-of-Thought Model}}}\vspace{0.1ex} \\
\textbf{IRCoT} 
& \textcolor{white}{0}505 &48.78\textcolor{white}{$^*$} & 62.36\textcolor{white}{$^*$} 
& \textcolor{white}{0}616 & 38.74\textcolor{white}{$^*$} & 47.29\textcolor{white}{$^*$}
& \textcolor{white}{0}639 & 22.84\textcolor{white}{$^*$} & 29.69\textcolor{white}{$^*$} 
\\ 
\cmidrule(lr){1-10}\noalign{\vskip 0.1ex}
\multicolumn{10}{l}{\textit{\textbf{Mistral Reader with Context from \OURS{}}}}\vspace{0.1ex} \\

\textbf{\OURS{}-Triple} 
&\textcolor{white}{0}167 &50.68{$^*$} &65.49{$^*$} 
& \textcolor{white}{0}170 & \textbf{46.30}{$^*$} & \textbf{55.49}{$^*$}
& \textcolor{white}{0}177 & \textbf{31.86}{$^*$} & \textbf{40.05}{$^*$}
\\ 
\textbf{\OURS{}-Doc} &\textcolor{white}{0}395 &\textbf{53.37}{$^*$} & \textbf{67.41}{$^*$} 
& \textcolor{white}{0}538 & 44.90{$^*$} & 54.32{$^*$}
& \textcolor{white}{0}508 & 30.66{$^*$} & 38.22{$^*$}
\\  
\bottomrule    
\end{tabular}
} 
\caption{Overall performance (\%) of \OURS{} and baselines on the test sets of three multi-hop QA datasets, where ``\# Tok'' is the average number of tokens in the documents used as context, $^*$ indicates p-value <0.05 compared with IRCoT. The best performance per dataset per metric is marked in boldface.}
\label{table:overall_performance_mistral_reader}
\end{table*}

\begin{table*}[tb]
\centering
\resizebox{0.9\textwidth}{!}{
\begin{tabular}{lcccccccccccc}
\toprule
\multirow{2}{*}{\textbf{Model}} & \multicolumn{3}{c}{\textbf{HotPotQA}} & \multicolumn{3}{c}{\textbf{2WikiMultiHopQA}} & \multicolumn{3}{c}{\textbf{MuSiQue}}  \\
\cmidrule(lr){2-4} \cmidrule(lr){5-7} \cmidrule(lr){8-10} 
& \textbf{\# Tok} & \textbf{EM} & \textbf{F1} & \textbf{\# Tok} & \textbf{EM} & \textbf{F1} & \textbf{\# Tok} & \textbf{EM} & \textbf{F1} \\
\midrule\noalign{\vskip 0.1ex}
\multicolumn{10}{l}{\textit{\textbf{Gemma Reader with Context from Naive Baselines}}}\vspace{0.1ex} \\
\textbf{w/o documents} 
& \textcolor{white}{00}43 & 20.95\textcolor{white}{$^*$} & 29.11\textcolor{white}{$^*$} 
& \textcolor{white}{00}40 &22.63\textcolor{white}{$^*$} & 26.60\textcolor{white}{$^*$}
& \textcolor{white}{00}44 & \textcolor{white}{0}4.05\textcolor{white}{$^*$} & \textcolor{white}{0}9.28\textcolor{white}{$^*$}
\\
\textbf{w. all documents} 
& 1,450 & 44.32\textcolor{white}{$^*$} & 58.81\textcolor{white}{$^*$} 
& 1,069 & 31.48\textcolor{white}{$^*$} & 40.05\textcolor{white}{$^*$}
& 2,613 & 21.10\textcolor{white}{$^*$} & 29.87\textcolor{white}{$^*$}
\\
\cmidrule(lr){1-10}\noalign{\vskip 0.1ex}
\multicolumn{10}{l}{\textit{\textbf{Gemma Reader with Context from Bi-Encoders}}}\vspace{0.1ex} \\
\textbf{Contriever} 
&1,450 &45.66\textcolor{white}{$^*$} &60.08\textcolor{white}{$^*$}  
& 1,069 & 33.18\textcolor{white}{$^*$} & 42.58\textcolor{white}{$^*$}
& 1,443 & 23.05\textcolor{white}{$^*$} & 31.76\textcolor{white}{$^*$}
\\
\textbf{DRAGON+} 
&1,450 &45.79\textcolor{white}{$^*$} &60.15\textcolor{white}{$^*$} 
& \textcolor{white}{0}446 & 32.59\textcolor{white}{$^*$} & 40.18\textcolor{white}{$^*$}
& 1,411 & 23.17\textcolor{white}{$^*$} & 31.74\textcolor{white}{$^*$}
\\
\textbf{E5} 
&\textcolor{white}{0}681 &45.46\textcolor{white}{$^*$} &59.32\textcolor{white}{$^*$} 
& \textcolor{white}{0}904 & 33.23\textcolor{white}{$^*$} & 42.55\textcolor{white}{$^*$}
& \textcolor{white}{0}705 & 23.21\textcolor{white}{$^*$} & 31.35\textcolor{white}{$^*$}
\\
\textbf{E5-Mistral} 
&\textcolor{white}{0}688 &47.47\textcolor{white}{$^*$} &61.83\textcolor{white}{$^*$}  
& \textcolor{white}{0}450 & 35.43\textcolor{white}{$^*$} & 43.72\textcolor{white}{$^*$}
& 1,112 & 25.94\textcolor{white}{$^*$} & 34.59\textcolor{white}{$^*$}
\\
\cmidrule(lr){1-10}\noalign{\vskip 0.1ex}
\multicolumn{10}{l}{\textit{\textbf{Gemma Reader with Context from Cross-Encoders}}}\vspace{0.1ex} \\
\textbf{monoT5} 
&1,450 &45.71\textcolor{white}{$^*$} &60.27\textcolor{white}{$^*$} 
& 1,069 & 32.73\textcolor{white}{$^*$} & 41.87\textcolor{white}{$^*$}
& 1,136 & 25.28\textcolor{white}{$^*$} & 33.81\textcolor{white}{$^*$}
\\
\textbf{BGE} 
&\textcolor{white}{0}709 &47.24\textcolor{white}{$^*$} &61.59\textcolor{white}{$^*$} 
& \textcolor{white}{0}683 & 34.62\textcolor{white}{$^*$} & 43.37\textcolor{white}{$^*$}
& \textcolor{white}{0}754 & 26.89\textcolor{white}{$^*$} & 35.54\textcolor{white}{$^*$}
\\
\textbf{RankLLaMA} 
&1,450 &45.59\textcolor{white}{$^*$} &59.84\textcolor{white}{$^*$} 
& \textcolor{white}{0}915 & 32.69\textcolor{white}{$^*$} & 41.45\textcolor{white}{$^*$}
& 1,476 & 25.90\textcolor{white}{$^*$} & 33.91\textcolor{white}{$^*$}
\\
\cmidrule(lr){1-10}\noalign{\vskip 0.1ex}
\multicolumn{10}{l}{\textit{\textbf{Gemma Reader with Context from Chain-of-Thought Model}}}\vspace{0.1ex} \\
\textbf{IRCoT}
&\textcolor{white}{0}452 &48.47\textcolor{white}{$^*$} &62.89\textcolor{white}{$^*$} 
& \textcolor{white}{0}551 & 38.11\textcolor{white}{$^*$} & 47.05\textcolor{white}{$^*$}
& \textcolor{white}{0}572 & 27.70\textcolor{white}{$^*$} & 36.27\textcolor{white}{$^*$}
\\ 
\cmidrule(lr){1-10}\noalign{\vskip 0.1ex}
\multicolumn{10}{l}{\textit{\textbf{Gemma Reader with Context from \OURS{}}}}\vspace{0.1ex} \\
\textbf{\OURS{}-Triple} 
&\textcolor{white}{0}147 & 51.68{$^*$} & \textbf{67.13}{$^*$} 
& \textcolor{white}{0}150 & \textbf{46.94}{$^*$} & \textbf{56.49}{$^*$}
& \textcolor{white}{0}157 & \textbf{34.17}{$^*$} & \textbf{42.49}{$^*$}
\\ 
\textbf{\OURS{}-Doc} 
&\textcolor{white}{0}353 & \textbf{51.80}{$^*$} & 66.81{$^*$} 
& \textcolor{white}{0}481 & 43.72{$^*$} & 53.70{$^*$}
& \textcolor{white}{0}455 & 33.06{$^*$} & 41.25{$^*$}
\\  
\bottomrule    
\end{tabular}
} 
\caption{Overall performance (\%) of \OURS{} (using Gemma-7B as reader) and baselines on the test sets of three multi-hop QA datasets, where ``\# Tok'' is the average number of tokens in the documents used as context, $^*$ indicates p-value <0.05 compared with IRCoT. The best performance per dataset per metric is marked in boldface.}
\label{table:overall_performance_gemma_reader}
\end{table*}

\begin{table}[tb]
\centering
\resizebox{0.48\textwidth}{!}{
\begin{tabular}{lcccccc}
\toprule
\multirow{2}{*}{\textbf{Model}} & \multicolumn{2}{c}{\textbf{HotPotQA}} & \multicolumn{2}{c}{\textbf{2WikiMultiHopQA}} & \multicolumn{2}{c}{\textbf{MuSiQue}}  \\
\cmidrule(lr){2-3} \cmidrule(lr){4-5} \cmidrule(lr){6-7} 
& \textbf{EM} & \textbf{F1} & \textcolor{white}{00}\textbf{EM} & \textbf{F1} & \textbf{EM} & \textbf{F1}  \\
\midrule

\textbf{\OURS{}-Triple} & \textbf{67.00}\textcolor{white}{$^*$} & \textbf{76.16}\textcolor{white}{$^*$} & \textcolor{white}{00}\textbf{66.40}\textcolor{white}{$^*$} & \textbf{72.52}\textcolor{white}{$^*$} & \textbf{40.40}\textcolor{white}{$^*$} & \textbf{46.92}\textcolor{white}{$^*$} \\

\cmidrule(lr){1-7}\noalign{\vskip 0.5ex}
\multicolumn{7}{l}{\textit{\textbf{Effectiveness of Using Demonstrations}}}\vspace{0.3em} \\ 
\textbf{w/o Demonstrations} & 63.20{$^*$} & 73.57{$^*$} & \textcolor{white}{00}64.40{$^*$} & 70.78{$^*$} & 38.20{$^*$} & 44.77{$^*$} \\

\cmidrule(lr){1-7}\noalign{\vskip 0.5ex}
\multicolumn{7}{l}{\textit{\textbf{Effectiveness of Using Adaptive Demonstrations}}}\vspace{0.3em} \\ 
\textbf{Fixed Demonstrations} &65.20$^*$ & 74.49$^*$ & \textcolor{white}{00}66.20\textcolor{white}{$^*$} & 71.85\textcolor{white}{$^*$} & 37.60$^*$ & 44.37$^*$ \\

\cmidrule(lr){1-7}\noalign{\vskip 0.5ex}
\multicolumn{7}{l}{\textit{\textbf{Efffects of the Number of Demonstrations}}}\vspace{0.3em} \\ 
\textbf{1 Demonstration} &65.00$^*$ & 74.76$^*$ & \textcolor{white}{00}67.40\textcolor{white}{$^*$} & 73.37\textcolor{white}{$^*$} & 38.00$^*$ & 44.84$^*$  \\
\textbf{5 Demonstrations} & 65.00$^*$ & 74.40$^*$ & \textcolor{white}{00}68.20$^*$ & 74.81$^*$ & 39.20\textcolor{white}{$^*$} & 46.41\textcolor{white}{$^*$} \\
\textbf{10 Demonstrations} & 64.60$^*$ & 74.31$^*$ & \textcolor{white}{00}65.01\textcolor{white}{$^*$} & 70.77\textcolor{white}{$^*$} & 37.65$^*$ & 43.33$^*$ \\

\bottomrule    
\end{tabular}
} 
\caption{Performance (\%) of \OURS{}-Triple under different variants of the reasoning chain constructor on the development sets of three QA datasets, where $^*$ denotes p<0.05 compared with \OURS{}-Triple.}
\label{table:ablation_study_demonstrations}
\end{table}

In order to examine the generalisation of reasoning chains, we report the performance of \OURS{} and baselines using different readers. Specifically, we leverage {\tt Mistral-7B-v0.1} and {\tt Gemma-7B} as the readers and report the corresponding performance in Table~\ref{table:overall_performance_mistral_reader} and Table~\ref{table:overall_performance_gemma_reader}, respectively. The results are similar to those obtained using LLaMA3 as the reader, demonstrating that the reasoning chains constructed by \OURS{} can effectively generalise across different readers. This consistency across various readers indicates the robustness of \OURS{} in producing reasoning chains that are not tailored to a specific model but are broadly applicable.

\begin{table*}[tb]
\centering
\begin{small}
\begin{tabular}{p{0.95\textwidth}}
\toprule
\textbf{Title}: Ellen Glasgow \vspace{0.2em} \\ 
\textbf{Text}: Ellen Anderson Gholson Glasgow (April 22, 1873 - 2013 November 21, 1945) was an American novelist who portrayed the changing world of the contemporary South. \vspace{0.2em} \\ 
\textbf{Knowledge Triples}: \vspace{0.2em} \\ 
<Ellen Glasgow; full name; Ellen Anderson Gholson Glasgow>, <Ellen Glasgow; date of birth; April 22, 1873>, <Ellen Glasgow; date of death; November 21, 1945>, <Ellen Glasgow; nationality; American>, <Ellen Glasgow; occupation; novelist>, <Ellen Glasgow; the theme of her literary work; changing world of the contemporary South> \\ 

\noalign{\vskip 1.0ex}\hdashline\noalign{\vskip 1.0ex}
\textbf{Title}: 
Heinrich von Bülow (Grotekop) \vspace{0.2em} \\ 
\textbf{Text}: 
Heinrich von Bülow also known as Big Top (Grotekop) was a knight born in the middle of the fourteenth century. He died either before 1395 or during 1415. He prospered as a warrior-supporter of Prince Albrecht of Mecklenburg (and of Sweden). Outside Mecklenberg, Heinrich Grotekop is still remembered in many quarters as an archetypal robber baron on account of his appetite for feuding. \vspace{0.2em} \\ 
\textbf{Knowledge Triples}: \vspace{0.2em} \\
<Heinrich von Bülow (Grotekop); also known as; Big Top (Grotekop)>, <Heinrich von Bülow (Grotekop); born in; middle of the fourteenth century>, <Heinrich von Bülow (Grotekop); died; before 1395 or during 1415>, <Heinrich von Bülow (Grotekop); occupation; warrior-supporter>, <Heinrich von Bülow (Grotekop); supported; Prince Albrecht of Mecklenburg (and of Sweden)>, <Heinrich von Bülow (Grotekop); remembered as; archetypal robber baron>, <Heinrich von Bülow (Grotekop); characterized by; appetite for feuding> \\ 

\noalign{\vskip 1.0ex}\hdashline\noalign{\vskip 1.0ex}
\textbf{Title}: 
Inaindha Kaigal \\ 
\textbf{Text}: 
Inaindha Kaigal (English: Conjoined Hands ) is a 1990 Indian Tamil film, directed by N. K. Vishwanathan. The film features C. Arunpandian, Ramki, Nirosha and Sindhu in lead roles, with Nassar, Senthil, Srividya, Murali Kumar and Prabhakaran playing supporting roles. The film, produced by Aabavanan who also wrote the script and lyrics, had musical score by Gyan Varma and was released on 2 August 1990. The film is a blockbuster in the year 1990 and became a successful venture. The film has been dubbed in Hindi as "Aakhri Sangam" and in Telugu as Sahasa Ghattam. \\ 
\textbf{Knowledge Triples}: \\
<Inaindha Kaigal; English translation; Conjoined Hands>, <Inaindha Kaigal; the year when the film was released; 1990>, <Inaindha Kaigal; genre of the film; Indian Tamil film>, <Inaindha Kaigal; director of the film; N. K. Vishwanathan>, <Inaindha Kaigal; lead actors; C. Arunpandian, Ramki, Nirosha, Sindhu>, <Inaindha Kaigal; supporting actors; Nassar, Senthil, Srividya, Murali Kumar, Prabhakaran>, <Inaindha Kaigal; individual who produced and also wrote the script and lyrics for the film; Aabavanan>, <Inaindha Kaigal; composer of the film's musical score; Gyan Varma>, <Inaindha Kaigal; release date of the film; 2 August 1990>, <Inaindha Kaigal; the status of the film in its release year; blockbuster in 1990 and became a successful venture>, <Inaindha Kaigal; Hindi version name of the film; Aakhri Sangam>, <Inaindha Kaigal; Telugu version name of the film; Sahasa Ghattam> \\ 
\bottomrule    
\end{tabular}
\end{small}
\caption{Examples of labelled data on HotPotQA dataset for the KG generator.}
\label{table:examples_for_kg_generator}
\end{table*}

\begin{table*}[tb]
\centering
\begin{small}
\begin{tabular}{p{0.95\textwidth}}
\toprule
\textbf{question}: Which magazine published papers more often; The Wittenburg Door or Sports Collectors Digest? \\ 
\textbf{reasoning chain}: <Sports Collectors Digest; type; American advertising weekly paper>, <The Wittenburg Door; publication frequency; bimonthly> \\
\textbf{Step-1}: \\ 
\textbf{existing knowledge triples}: \\ 
\textbf{candidate knowledge triples}: \\ 
A. no need for additional knowledge triples  \\
B. <Sports Collectors Digest; purpose; provides an avenue through which sellers, traders and avid buyers of sports memorabilia may interact> \\ 
C. <The Wittenburg Door; type; Christian satire and humor magazine> \\ 
D. <Mike Yaconelli; role in The Wittenburg Door; satirical magazine> \\ 
E. <Sports Collectors Digest; type; American advertising weekly paper> \\ 
\textbf{the next possible triple is}: E \\ 

\textbf{Step-2}: \\ 
\textbf{existing knowledge triples}: <Sports Collectors Digest; type; American advertising weekly paper> \\ 
\textbf{candidate knowledge triples}: \\ 
A. no need for additional knowledge triples  \\
B. <The Wittenburg Door; type; Christian satire and humor magazine> \\ 
C. <Mike Yaconelli; role in The Wittenburg Door; satirical magazine> \\ 
D. <The Wittenburg Door; publication frequency; bimonthly> \\ 
E.  <The Wittenburg Door; start year of publication; 1971> \\ 
\textbf{the next possible triple is}: D \\

\textbf{Step-3}: \\ 
\textbf{existing knowledge triples}: <Sports Collectors Digest; type; American advertising weekly paper>, <The Wittenburg Door; publication frequency; bimonthly> \\ 
\textbf{candidate knowledge triples}: \\ 
A. no need for additional knowledge triples  \\
B. <Sports Collectors Digest; purpose; provides an avenue through which sellers, traders and avid buyers of sports memorabilia may interact> \\ 
C. <The Wittenburg Door; type; Christian satire and humor magazine> \\ 
D. <Mike Yaconelli; role in The Wittenburg Door; satirical magazine> \\ 
E. <The Wittenburg Door; reference to; the door of the All Saints' Church in Wittenberg> \\ 
\textbf{the next possible triple is}: A \\ 

\noalign{\vskip 1.0ex}\hdashline\noalign{\vskip 1.0ex}

\textbf{question}: What is the 5th studio album released by the singer of "A Girl's Gotta Do (What a Girl's Gotta Do)"? \\ 
\textbf{reasoning chain}: \\ 
\textbf{Step-1}: \\ 
\textbf{existing knowledge triples}: \\
\textbf{candidate knowledge triples}: \\ 
A. no need for additional knowledge triples \\ 
B. <A Girl's Gotta Do (What a Girl's Gotta Do); artist; American country music artist Mindy McCready> \\ 
C. <A Girl's Gotta Do (What a Girl's Gotta Do); release date; February 1997> \\ 
D. <Ten Thousand Angels; fourth single; "A Girl's Gotta Do (What a Girl's Gotta Do)"> \\ 
E. <A Girl's Gotta Do (What a Girl's Gotta Do); songwriters; Robert Byrne and Rick Bowles> \\
\textbf{the next possible triple is}: B

\textbf{Step-2}: \\ 
\textbf{existing knowledge triples}: <A Girl's Gotta Do (What a Girl's Gotta Do); artist; American country music artist Mindy McCready> \\
\textbf{candidate knowledge triples}:  \\ 
A. no need for additional knowledge triples \\ 
B. <Mindy McCready; number of studio albums; five> \\ 
C. <Mindy McCready; fourth album; self-titled> \\ 
D. <Mindy McCready; debut album release year; 1996> \\ 
E. <Mindy McCready; fifth album; I'm Still Here> \\
\textbf{the next possible triple is}: E 

\textbf{Step-3}: \\ 
\textbf{existing knowledge triples}: <A Girl's Gotta Do (What a Girl's Gotta Do); artist; American country music artist Mindy McCready>, <Mindy McCready; fifth album; I'm Still Here> \\
\textbf{candidate knowledge triples}: \\ 
A. no need for additional knowledge triples \\ 
B. <Mindy McCready; third album; I'm Not So Tough> \\ 
C. <Mindy McCready; debut album release year; 1996> \\ 
D. <Mindy McCready; fourth album; self-titled> \\
E. <Mindy McCready; number of studio albums; five> \\ 
\textbf{the next possible triple is}: A \\

\bottomrule    
\end{tabular}
\end{small}
\vspace{-0.5em}
\caption{Examples of labelled data on HotPotQA dataset for the triple selector.}
\label{table:examples_for_triple_selector}
\vspace{-0.5em}
\end{table*}

\subsection{Ablation Studies on \OURS{}-Doc}
\label{app:ablation_of_trace_doc}
In the main text, we provide ablation study results for \OURS{}-Triple to verify the effectiveness of each component in \OURS{}, i.e., KG generator, reasoning chain constructor, triple ranker and triple selector. 
We further conduct ablation studies on \OURS{}-Doc to investigate \jy{the impacts of these components on the effectiveness of using reasoning chains to retrieve relevant documents from $\mathcal{D}_q$.} 
These ablation studies are conducted in a manner similar to that of \OURS{}-Triple, as described in the main text. 
The experimental results, presented in Table~\ref{table:ablation_study_use_document}, demonstrate consistent findings with the results of \OURS{}-Triple, highlighting the importance and effectiveness of each component in the overall performance of \OURS{}. 

\begin{figure}[tb]
\begin{center}
\includegraphics[width=0.3\textwidth]{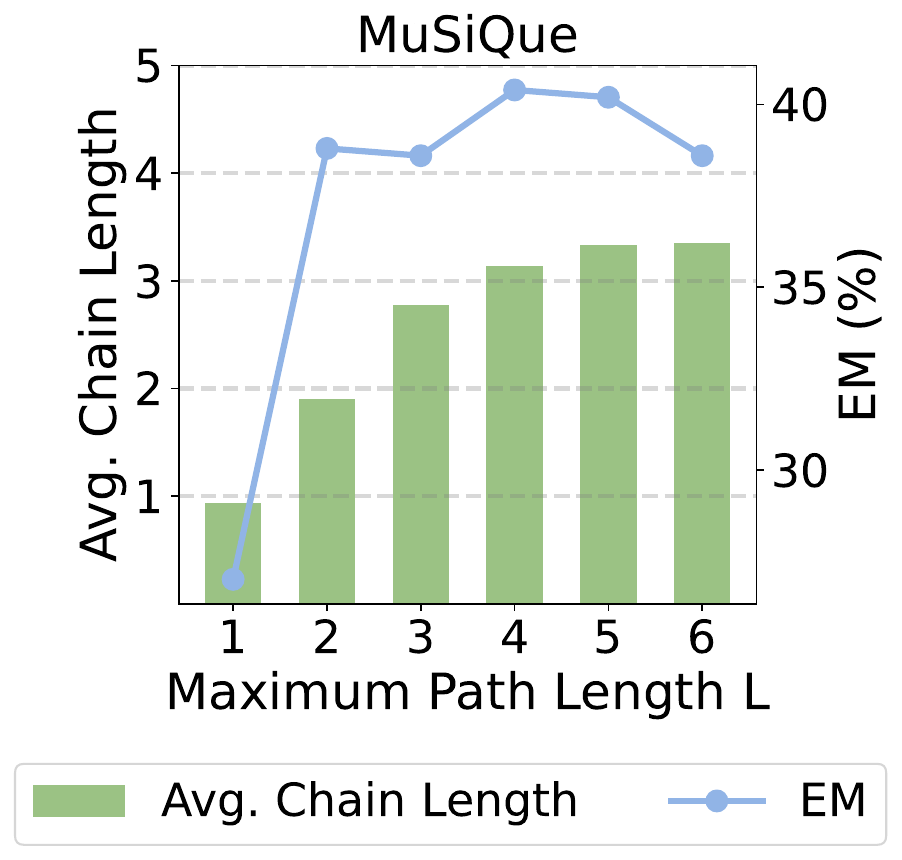}
\end{center}
\caption{QA performance (\%) and average chain length of \OURS{}-Triple under different values of $L$ on the development set of MuSiQue dataset.}
\label{figure:maximum_path_length_part3}
\end{figure}

\subsection{Effects of Maximum Chain Length L on MuSiQue}
\label{app:effects_of_l_on_musique}
The experimental results regarding the effects of $L$ on the MuSiQue dataset are presented in Figure~\ref{figure:maximum_path_length_part3}. These results are consistent with those observed on the HotPotQA and 2WikiMultiHopQA datasets, demonstrating the effectiveness of the adaptive chain termination strategy. The results also highlight the importance of setting a proper maximum reasoning chain length $L$ to achieve optimal performance.

\subsection{Effects of Demonstrations in Reasoning Chain Constructor}
\label{app:effects_of_demonstrations}
In the reasoning chain constructor of \OURS{}, we include demonstrations in the prompt to guide the construction of reasoning chains. We conduct ablation studies to investigate the effects of the demonstrations on the overall performance of \OURS{}. 

Specifically, to examine the effectiveness of the demonstrations, we introduce a variant of \OURS{}-Triple, namely \textit{w/o demonstrations}, where we remove the demonstrations and only use the instruction to guide the reasoning chain construction. The results presented in Table~\ref{table:ablation_study_demonstrations} show that removing the demonstrations significantly degrades the performance of \OURS{}-Triple on all the datasets. This indicates the effectiveness of the demonstrations in enhancing the reasoning chain construction and overall performance of \OURS{}-Triple. 

Moreover, as described in Appendix~\ref{app:implementation}, we use E5-Mistral to retrieve the top three most similar questions and their corresponding reasoning chains as demonstrations when constructing reasoning chains for a given question. To examine the effectiveness of such an adaptive demonstration selection approach, we introduce a variant of \OURS{}-Triple, namely \textit{Fixed Demonstrations}, where the same set of demonstrations is used for all the questions. The results, provided in Table~\ref{table:ablation_study_demonstrations}, indicate that using fixed demonstrations significantly degrades the performance on the HotPotQA and MuSiQue datasets, and slightly degrades the performance on the 2WikiMultiHopQA dataset. These findings demonstrate the effectiveness of the adaptive demonstration selection approach, which is also consistent with previous works~\cite{rubin2022learning,li2023unified}. 

Furthermore, we conduct experiments to investigate the effects of the number of demonstrations on the overall performance of \OURS{}-Triple. Specifically, we vary the number of demonstrations to $1, 3, 5, 10$ and report the corresponding performance of \OURS{}-Triple. Note that \OURS{}-Triple uses 3 demonstrations by default. The results are presented in Table~\ref{table:ablation_study_demonstrations}, which indicate that increasing the number of demonstrations does not necessarily improve the performance. For example, the performance of \OURS{}-Triple with $10$ demonstrations is generally worse than with fewer demonstrations. This might due to the fact that adding more demonstrations can introduce irrelevant data, which may distract the reasoning chain constructor and lead to incorrect decisions.

\subsection{Effects of the Number of Reasoning Chains R}
\label{app:effects_of_R}

\begin{figure}[tb]
\begin{center}
\includegraphics[width=0.48\textwidth]{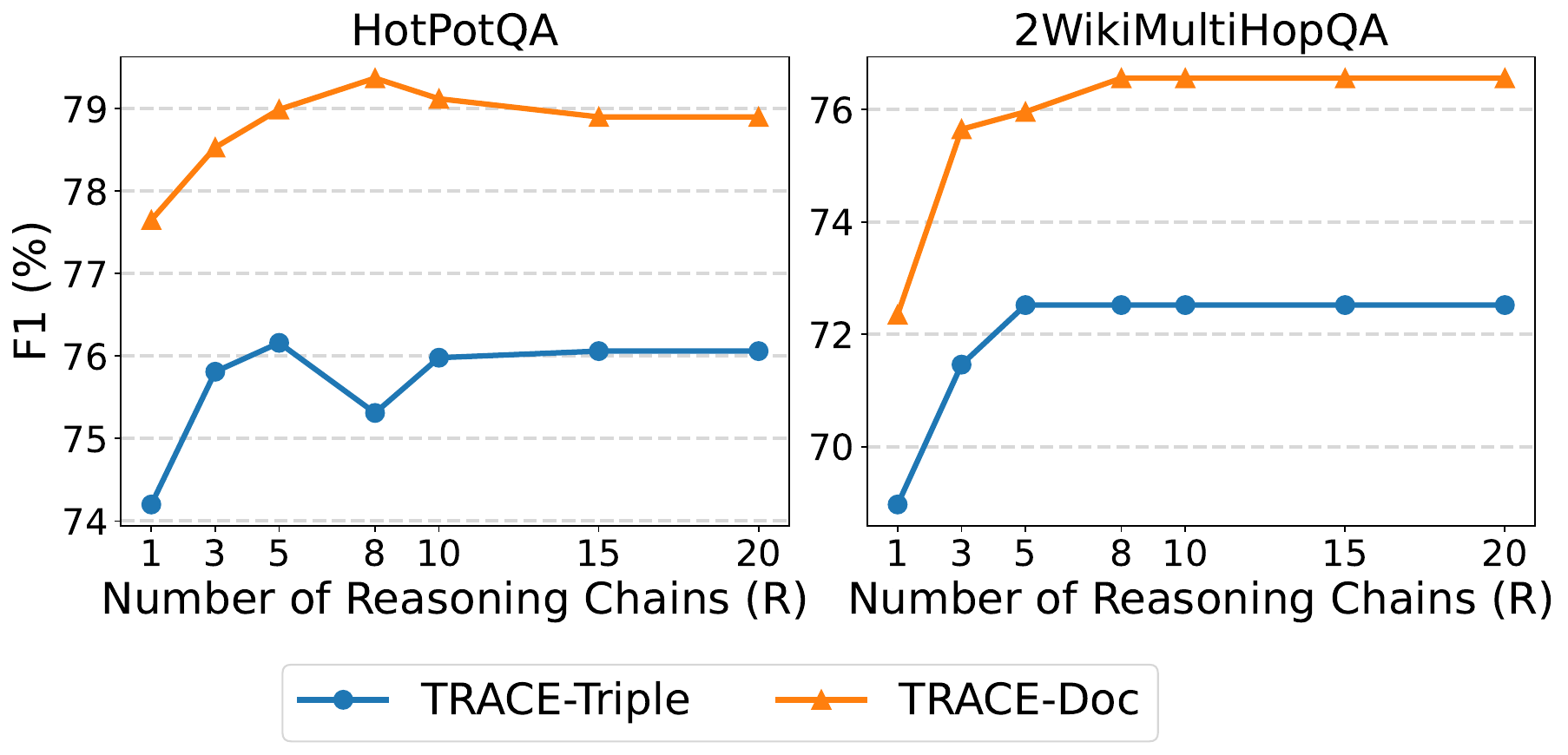}
\end{center}
\caption{Performance (\%) of \OURS{} with different numbers of reasoning chains on the development sets of the HotPotQA and 2WikiMultiHopQA datasets.}
\label{figure:effect_of_R}
\end{figure}

\begin{figure}[tb]
\begin{center}
\includegraphics[width=0.48\textwidth]{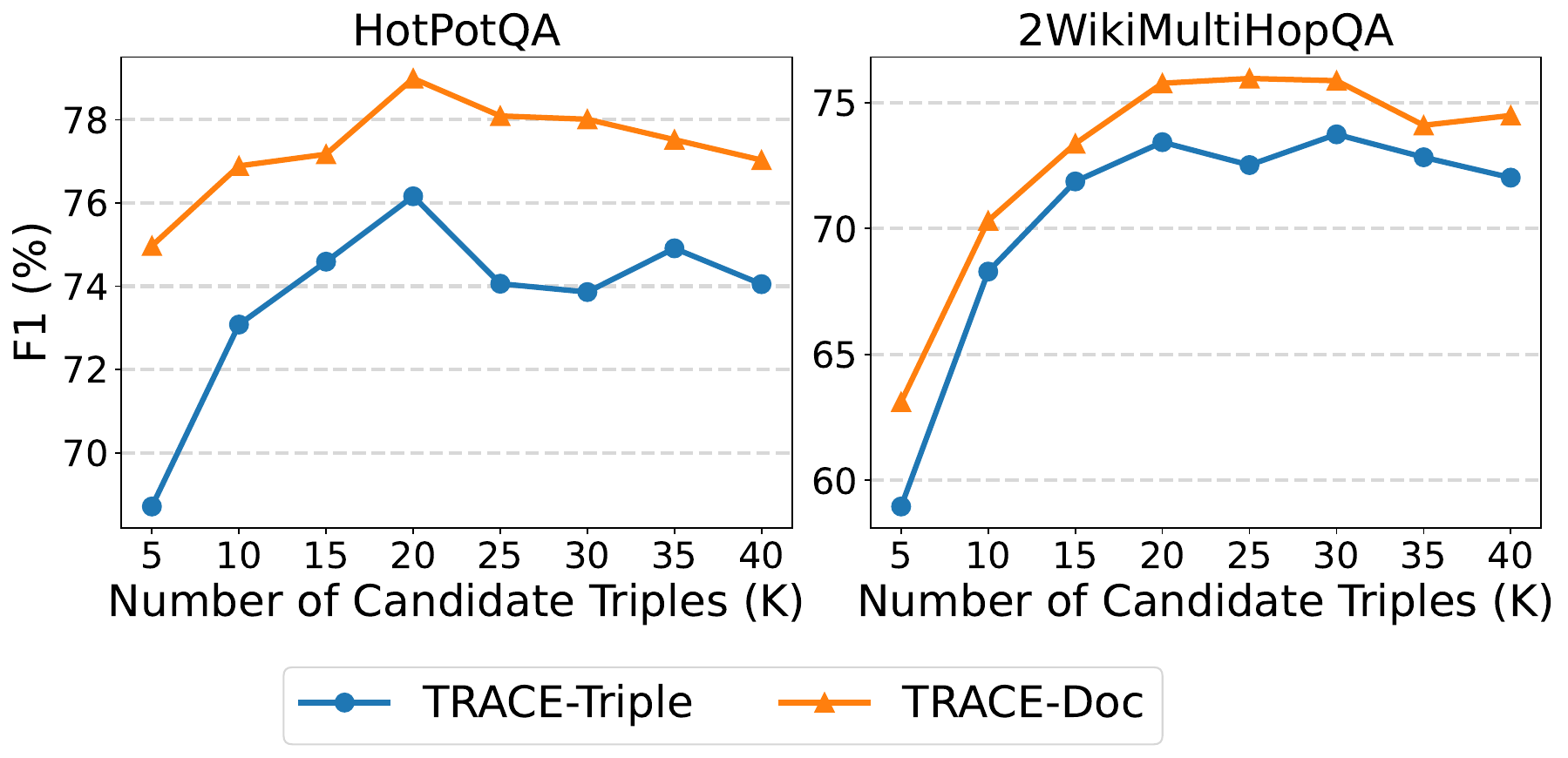}
\end{center}
\caption{Performance (\%) of \OURS{} with different numbers of candidate triples on the development sets of the HotPotQA and 2WikiMultiHopQA datasets.}
\label{figure:effect_of_K}
\end{figure}

\begin{figure}[tb]
\begin{center}
\includegraphics[width=0.48\textwidth]{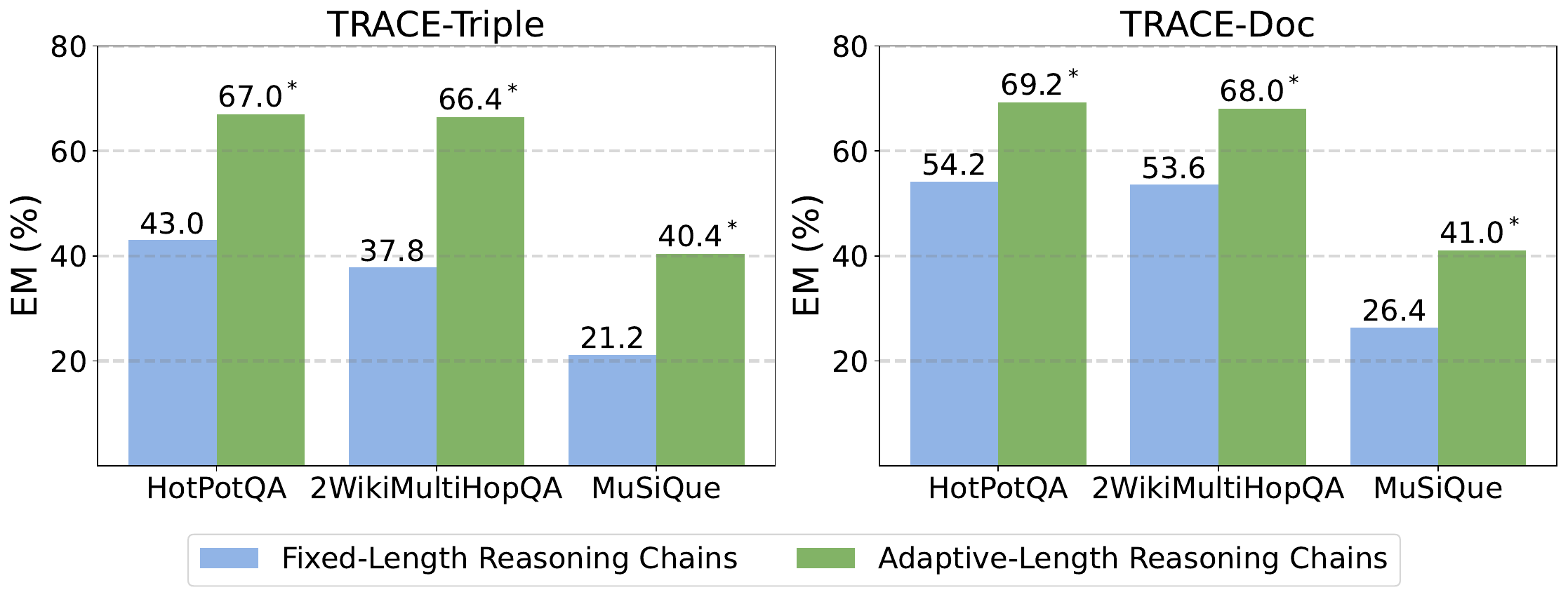}
\end{center}
\caption{Performance (\%) of \OURS{} under different reasoning chain settings on the development sets of three multi-hop QA datasets, where $^*$ indicates p-value<0.05.}
\label{figure:effect_of_adaptive_reasoning_paths}
\vspace{-0.5em}
\end{figure}

To investigate the effects of the number of reasoning chains $R$, we vary the value of $R$ from 1 to 20 and report the corresponding performance. Figure~\ref{figure:effect_of_R} shows the performance of both \OURS{}-Triple and \OURS{}-Doc on the development sets of HotPotQA and 2WikiMultiHopQA datasets. The results indicate that as $R$ increases, the performance of both \OURS{}-Triple and \OURS{}-Doc initially improves and then becomes stable. This trend can be explained by the fact that initially increasing the number of reasoning chains helps to incorporate additional information that is missing in the previous reasoning chains. However, once a certain threshold is reached, no further relevant information can be added, leading to performance stabilisation. 

\subsection{Effects of the Number of Candidate Triples K}
\label{app:effects_of_K}

In the reasoning chain constructor of \OURS{}, a triple ranker is used to select a subset of candidate triples from the KG. To investigate the effects of the number of candidate triples $K$, we vary the value of $K$ from 5 to 40 and report the corresponding performance. The results are reported in Figure~\ref{figure:effect_of_K}, which shows the performance of \OURS{}-Triple and \OURS{}-Doc on the HotPotQA and 2WikiMultiHopQA datasets. These results indicate that as $K$ increases, the performance of both \OURS{}-Triple and \OURS{}-Doc initially improves but then declines after a certain threshold, such as $20$ for HotPotQA and $30$ for 2WikiMultiHopQA. This can be explained by the balance between information richness and noise. Initially, increasing $K$ allows for the inclusion of potentially relevant triples, thereby improving the performance. However, beyond a certain threshold, using too many triples can introduce noise and irrelevant information, which can distract the triple selector and degrade the overall performance. 

\subsection{Effects of Adaptive Chain Termination}
\label{app:adaptive_chain_termination}

In the reasoning chain constructor of \OURS{}, an adaptive chain termination strategy is employed to automatically determine the optimal lengths of reasoning chains. We conduct ablation studies to investigate the effectiveness of this approach. Specifically, we introduce a variant of \OURS{}, namely \textit{Fixed-Length Reasoning Chains}, where the lengths of all the reasoning chains are set to a fixed value $L$ (4 in our experiments). The results are presented in Figure~\ref{figure:effect_of_adaptive_reasoning_paths}, which shows the QA performance of \OURS{}-Triple and \OURS{}-Doc on three multi-hop QA datasets. The results indicate that \OURS{} with adaptive-length reasoning chains (both \OURS{}-Triple and \OURS{}-Doc) significantly outperforms its fixed-length counterpart by a large margin. This is because the adaptive chain termination strategy allows the reasoning chain constructor to dynamically determine the required number of triples, thereby avoiding the introduction of unnecessary and redundant triples and leading to the improved performance.

\subsection{Case Study}
\label{app:case_study}

The complete results of the case study are provided in Table~\ref{table:case_study_full_results}, which shows the top-5 reasoning chains obtained with the reasoning chain constructor and the resulting relevant documents identified using these reasoning chains. These cases yield the following additional findings: 

\noindent (1) Generating multiple reasoning chains is beneficial as it helps to incorporate relevant information that may be missing in the initial reasoning chains. For example, in the reasoning chains for the question ``\textit{What is the birth date of this Spanish footballer, who was added as a holding midfielder in the 2021-13 FC Bayern Munich season?}'', the KG triples ``$\langle$\textit{Javi Martínez; position; defensive midfielder or a central defender}$\rangle$'' in the second reasoning chain and ``$\langle$\textit{Javi Martínez; nationality; Spanish}$\rangle$'' in the third reasoning chain provide complementary information to the first reasoning chain to correctly answer the question. Therefore, using multiple reasoning chains helps to provide a more enriched context, leading to improved performance. This finding is also consistent with the empirical results in presented in Appendix~\ref{app:effects_of_R}. 

\noindent (2) It is effective to leverage reasoning chains to identify supporting documents. For example, for the multi-hop question ``\textit{Are both Blaise Cendrars and Julian Barnes are a citizen of the same country?}'', only two documents are identified using the reasoning chains. The first document is about \textit{Blaise Cendrars} and the second one is about \textit{Julian Barnes}, both of which are relevant to the question and are highly useful to correctly answer the question. Similar results are also observed in the other two examples. Therefore, these findings indicate that using reasoning chains helps to identify relevant documents while avoiding the introduction of noisy documents, leading to the enhanced performance.

\begin{table*}[tb]
\centering
\begin{small}
\begin{tabular}{p{0.95\textwidth}}
\toprule
\textbf{Title}: Ellen Glasgow  \vspace{0.2em} \\
\textbf{Text}: Ellen Anderson Gholson Glasgow (April 22, 1873 – November 21, 1945) was an American novelist who portrayed the changing world of the contemporary South. \vspace{0.2em} \\ 
\textbf{Generated Knowledge Triples}: \vspace{0.2em} \\ 
<Ellen Glasgow; full name; Ellen Anderson Gholson Glasgow>, <Ellen Glasgow; date of birth; April 22, 1873>, <Ellen Glasgow; date of death; November 21, 1945>, <Ellen Glasgow; nationality; American>, <Ellen Glasgow; occupation; novelist>, <Ellen Glasgow; the theme of her literary work; changing world of the contemporary South> \\ 

\noalign{\vskip 1.0ex}\hdashline\noalign{\vskip 1.0ex}

\textbf{Title}: Julian Barnes \vspace{0.2em} \\ 
\textbf{Text}: Julian Patrick Barnes (born 19 January 1946) is an English writer. Barnes won the Man Booker Prize for his book "The Sense of an Ending" (2011), and three of his earlier books had been shortlisted for the Booker Prize: "Flaubert's Parrot" (1984), "England, England" (1998), and "Arthur \& George" (2005). He has also written crime fiction under the pseudonym Dan Kavanagh. In addition to novels, Barnes has published collections of essays and short stories. \vspace{0.2em} \\ 
\textbf{Generated Knowledge Triples}: \vspace{0.2em} \\ 
<Julian Barnes; nationality; English>, <Julian Barnes; date of birth; 19 January 1946>, <Julian Barnes; occupation; writer>, <Julian Barnes; award won; Man Booker Prize for "The Sense of an Ending" (2011)>, <Julian Barnes; books shortlisted for the Booker Prize; "Flaubert's Parrot" (1984), "England, England" (1998), "Arthur \& George" (2005)>, <Julian Barnes; pseudonym; Dan Kavanagh>, <Julian Barnes; genre; crime fiction>, <Julian Barnes; type of writing; novels, essays, short stories> \\

\noalign{\vskip 1.0ex}\hdashline\noalign{\vskip 1.0ex}

\textbf{Title}: Emarosa \vspace{0.2em} \\ 
\textbf{Text}: Emarosa ( ) is an American post-hardcore band from Lexington, Kentucky. The band currently consists of founding members ER White (lead guitar) and Jordan Stewart (keyboards), as well as lead vocalist Bradley Walden and rhythm guitarist Marcellus Wallace. \vspace{0.2em} \\ 
\textbf{Generated Knowledge Triples}: \vspace{0.2em} \\ 
<Emarosa; genre; post-hardcore>, <Emarosa; location; Lexington, Kentucky>, <Emarosa; members; ER White (lead guitar), Jordan Stewart (keyboards), Bradley Walden (lead vocalist), Marcellus Wallace (rhythm guitarist)> \\

\noalign{\vskip 1.0ex}\hdashline\noalign{\vskip 1.0ex}

\textbf{Title}: Tantalizers \vspace{0.2em} \\ 
\textbf{Text}: Tantalizers is a leading Nigerian fast food restaurant chain. It opened its first location c. 1997 Festac Town, Lagos. This first location was initially a small neighborhood restaurant serving hamburgers. Success at this first location led to an expansion that has seen the company and its franchisees open additional locations in cities such as Lagos, Ibadan, Abuja, and Port Harcourt. As of 2015, the restaurant has 50 outlets across Nigeria. \vspace{0.2em} \\ 
\textbf{Generated Knowledge Triples}: \vspace{0.2em} \\ 
<Tantalizers; type; fast food restaurant chain>, <Tantalizers; location; Festac Town, Lagos>, <Tantalizers; year of opening; c. 1997>, <Tantalizers; initial location; small neighborhood restaurant>, <Tantalizers; initial menu item; hamburgers>, <Tantalizers; expansion; additional locations in cities such as Lagos, Ibadan, Abuja, and Port Harcourt>, <Tantalizers; number of outlets; 50>, <Tantalizers; location of outlets; across Nigeria> \\ 

\noalign{\vskip 1.0ex}\hdashline\noalign{\vskip 1.0ex}

\textbf{Title}: Julius Caesar Chappelle \vspace{0.2em} \\ 
\textbf{Text}: Julius Caesar Chappelle (1852–1904) was an African-American politician born into slavery in South Carolina. After the American Civil War, he lived for a time with his family in LaVilla, Florida, helping develop the new town. In 1870 he was one of numerous Southern black migrants to Boston, Massachusetts, which had a thriving black community and strong abolitionist history. He later joined the Republican Party that was founded by abolitionists, and Chappelle was elected to two terms in the Massachusetts state legislature, serving 1883-1886. Julius Caesar Chappelle was also the first African-American to serve on the Massachusetts State Senate Committee where he served three terms. Chappelle was active in supporting civil rights, trying to reduce discrimination, and consumer affairs. His speeches were frequently covered by newspapers. Throughout his life and political career, he held secondary supervisory government positions in maintenance, such as at the United States Post Office and US Boston Custom House. Although Julius Caesar Chappelle may have graced the same pages in newspapers as Frederick Douglass, Chappelle is not as well-known because he is not known to have left much of a literary footprint such as writing manuscripts or for pamphlets. \vspace{0.2em} \\ 
\textbf{Generated Knowledge Triples}: \vspace{0.2em} \\ 
<Julius Caesar Chappelle; date of birth; 1852>, <Julius Caesar Chappelle; date of death; 1904>, <Julius Caesar Chappelle; place of birth; South Carolina>, <Julius Caesar Chappelle; nationality; African-American>, <Julius Caesar Chappelle; occupation; politician>, <Julius Caesar Chappelle; place of residence; LaVilla, Florida>, <Julius Caesar Chappelle; place of residence; Boston, Massachusetts>, <Julius Caesar Chappelle; political party; Republican Party>, <Julius Caesar Chappelle; served in; Massachusetts state legislature, 1883-1886>, <Julius Caesar Chappelle; served in; Massachusetts State Senate Committee, three terms>, <Julius Caesar Chappelle; role in civil rights; active in supporting civil rights, trying to reduce discrimination>, <Julius Caesar Chappelle; role in consumer affairs; active in consumer affairs>, <Julius Caesar Chappelle; occupation in government; held secondary supervisory government positions in maintenance, such as at the United States Post Office and US Boston Custom House>, <Julius Caesar Chappelle; notable achievement; first African-American to serve on the Massachusetts State Senate Committee>, <Julius Caesar Chappelle; notable achievement; served two terms in the Massachusetts state legislature>, <Julius Caesar Chappelle; notable achievement; served three terms in the Massachusetts State Senate Committee>, <Julius Caesar Chappelle; notable achievement; his speeches were frequently covered by newspapers> \\ 

\bottomrule    
\end{tabular}
\end{small}
\vspace{-0.5em}
\caption{Examples of generated KGs for documents on the HotPotQA dataset.}
\label{table:examples_of_generated_kgs}
\vspace{-0.5em}
\end{table*}

\begin{table*}[tb]
\centering
\begin{small}
\begin{tabular}{p{0.95\textwidth}}
\toprule
\textbf{Question}: Are both Blaise Cendrars and Julian Barnes are a citizen of the same country? \vspace{0.2em}\\
\textbf{Reasoning Chains}: \vspace{0.2em}\\
\textbf{1}: \textcolor{blue}{<Blaise Cendrars; nationality; Swiss>, <Julian Barnes; nationality; English>} \vspace{0.1em}\\
\textbf{2}: <Blaise Cendrars; nationality; French>, <Julian Barnes; nationality; English> \\ 
\textbf{3}: <Julian Barnes; nationality; English>, <Blaise Cendrars; nationality; French>, <Blaise Cendrars; nationality; Swiss>, <Blaise Cendrars; event; became a naturalized French citizen in 1916> \vspace{0.1em}\\
\textbf{4}: <Julian Barnes; nationality; English>, <Blaise Cendrars; nationality; Swiss>\\ 
\textbf{5}: <Blaise Cendrars; nationality; French>, <Julian Barnes; nationality; English>, <Blaise Cendrars; event; became a naturalized French citizen in 1916> \vspace{0.1em}\\
\textbf{Relevant Documents Identified Using Reasoning Chains}: \vspace{0.2em}\\ 
\textbf{1.} \textbf{Title}: Blaise Cendrars \vspace{0.2em}\\ 
Frédéric-Louis Sauser (1 September 1887 – 21 January 1961), better known as \textcolor{orange}{Blaise Cendrars}, was a \textcolor{orange}{Swiss-born} novelist and poet who \textcolor{orange}{became a naturalized French citizen} in 1916. He was a writer of considerable influence in the European modernist movement. \vspace{0.1em}\\ 
\textbf{2.} \textbf{Title}: Julian Barnes \vspace{0.1em}\\
\textcolor{orange}{Julian Patrick Barnes} (born 19 January 1946) is an \textcolor{orange}{English writer}. Barnes won the Man Booker Prize for his book "The Sense of an Ending" (2011), and three of his earlier books had been shortlisted for the Booker Prize: "Flaubert's Parrot" (1984), "England, England" (1998), and "Arthur \& George" (2005). He has also written crime fiction under the pseudonym Dan Kavanagh. In addition to novels, Barnes has published collections of essays and short stories. \vspace{0.1em}\\ 
\textbf{Generated Answer}: no \vspace{0.2em}\\

\noalign{\vskip 1.0ex}\hdashline\noalign{\vskip 1.0ex}
\textbf{Question}: What was the occupation of both Christina Stead and Nuruddin Farah? \vspace{0.2em}\\
\textbf{Reasoning Chains}: \vspace{0.2em}\\
\textbf{1}: \textcolor{blue}{<Christina Stead; occupation; novelist and short-story writer>, <Nuruddin Farah; occupation; novelist>} \vspace{0.1em}\\
\textbf{2}: <Christina Stead; occupation; novelist and short-story writer>, <Nuruddin Farah; written works; plays, short stories, essays> \\ 
\textbf{3}: <Nuruddin Farah; occupation; novelist>, <Christina Stead; occupation; novelist and short-story writer> \vspace{0.1em}\\
\textbf{4}: <Nuruddin Farah; birthdate; 24 November 1945>, <Christina Stead; occupation; novelist and short-story writer> \\ 
\textbf{5}: <Nuruddin Farah; written works; plays, short stories, essays>, <Christina Stead; occupation; novelist and short-story writer> \vspace{0.1em}\\
\textbf{Relevant Documents Identified Using Reasoning Chains}: \vspace{0.2em}\\ 
\textbf{1.} \textbf{Title}: Christina Stead \vspace{0.2em}\\ 
\textcolor{orange}{Christina Stead} (17 July 190231 March 1983) \textcolor{orange}{was an Australian novelist and short-story writer} acclaimed for her satirical wit and penetrating psychological characterisations. Christina Stead was a committed Marxist, although she was never a member of the Communist Party. She spent much of her life outside Australia. \vspace{0.1em}\\ 
\textbf{2.} \textbf{Title}: Nuruddin Farah \vspace{0.1em}\\
\textcolor{orange}{Nuruddin Farah} (Somali: "Nuuradiin Faarax") (born 24 November 1945) is a \textcolor{orange}{Somali novelist}. He has also written plays both for stage and radio, as well as short stories and essays. Since leaving Somalia in the 1970s he has lived and taught in numerous countries, including the United States, England, Germany, Italy, Sweden, Sudan, India, Uganda, Nigeria and South Africa.\vspace{0.1em}\\ 
\textbf{Generated Answer}: novelist \vspace{0.2em}\\

\noalign{\vskip 1.0ex}\hdashline\noalign{\vskip 1.0ex}

\textbf{Question}: What is the birth date of this Spanish footballer, who was added as a holding midfielder in the 2012-13 FC Bayern Munich season?  \vspace{0.2em}\\
\textbf{Reasoning Chains}: \vspace{0.2em}\\
\textbf{1}: \textcolor{blue}{<2012–13 FC Bayern Munich season; new player signed after the first week of the Bundesliga season; Javi Martínez>, <Javi Martínez; date of birth; 2 September 1988>} \vspace{0.1em}\\
\textbf{2}: \textcolor{blue}{<Javi Martínez; position; defensive midfielder or a central defender>}, <Javi Martínez; date of birth; 2 September 1988> \\ 
\textbf{3}: \textcolor{blue}{<Javi Martínez; nationality; Spanish>}, <Javi Martínez; date of birth; 2 September 1988> \vspace{0.1em}\\
\textbf{4}: <Javi Martínez; date of birth; 2 September 1988>, <Javi Martínez; club; FC Bayern Munich> \\ 
\textbf{5}: <Javi Martínez; club; FC Bayern Munich>, <Javi Martínez; date of birth; 2 September 1988> \vspace{0.1em}\\
\textbf{Relevant Documents Identified Using Reasoning Chains}: \vspace{0.2em}\\ 
\textbf{1.} \textbf{Title}: Javi Martínez \vspace{0.2em}\\ 
\textcolor{orange}{Javier "Javi" Martínez Aginaga} ( \textcolor{orange}{born 2 September 1988}) is a Spanish footballer who plays for German club FC Bayern Munich as a defensive midfielder or a central defender. \vspace{0.1em}\\ 
\textbf{2.} \textbf{Title}: 2012–13 FC Bayern Munich season \vspace{0.1em}\\
\textcolor{orange}{The 2012–13 FC Bayern Munich season} was the 114th season in the club's history and the 48th consecutive season in the top flight of German football, the Bundesliga, since the promotion of the team from the Regionalliga Süd in 1965. Before the start of the season, Bayern signed Xherdan Shaqiri, Dante, Claudio Pizarro, Mitchell Weiser, Tom Starke and Mario Mandžukić. \textcolor{orange}{Bayern also added holding midfielder Javi Martínez after the first week of the Bundesliga season at the transfer deadline.} The club started the season with a nine-match winning streak. The club would end the season claiming the Treble, winning the Bundesliga, the UEFA Champions League and the DFB-Pokal. 
\vspace{0.1em}\\ 
\textbf{Generated Answer}: 2 September 1988 \vspace{0.2em}\\

\bottomrule    
\end{tabular}
\end{small}
\vspace{-0.5em}
\caption{Case study of \OURS{} on the HotPotQA dataset.}
\label{table:case_study_full_results}
\vspace{-0.5em}
\end{table*}

\end{document}